%% file: sample-acmlarge.tex
\documentclass[acmlarge, nonacm]{acmart}

\AtBeginDocument{%
  }

\usepackage{graphicx}
\usepackage{subcaption}
\usepackage{float}
\usepackage{makecell}


\begin{document}

\title[Restoring Ancient Inscription Textures]{MESA: A Training‑Free Multi‑Exemplar Deep Framework for Restoring Ancient Inscription Textures}
\author{Vasilis Toulatzis}
\email{vtoulatz@cse.uoi.gr}
\author{Sofia Theodoridou}
\email{s.theo@outlook.com}
\author{Ioannis Fudos}
\email{fudos@cse.uoi.gr}
\affiliation{%
  \department{Department of Computer Science and Engineering}
  \institution{University of Ioannina}
 \city{Ioannina}
 \country{Greece}
}

\begin{teaserfigure}
  \centering
  \includegraphics[width=\textwidth]{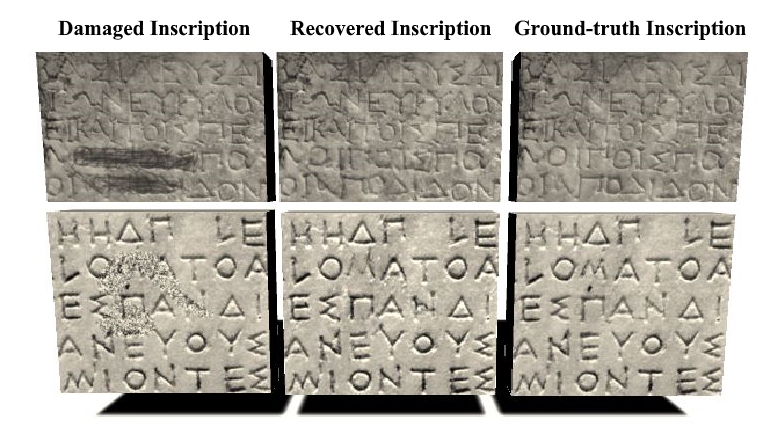}
  \caption{Restoring Ancient Inscription Textures: overview of our MESA training free approach. On the left is the inscription with text partially damaged, in the middle is the restored texture revealing the damaged letters and, on the right is the inscription used as ground-truth texture}
  \Description{High-level overview of the proposed method.}
\end{teaserfigure}

\begin{abstract}

Ancient inscriptions frequently suffer missing or corrupted regions from fragmentation, erosion, or other damage, hindering reading, and analysis. We review prior image restoration methods and their applicability to inscription image recovery, then introduce MESA (Multi-Exemplar, Style-Aware) — an image-level restoration method that uses well-preserved exemplar inscriptions (from the same epigraphic monument, material, or similar letterforms) to guide reconstruction of damaged text. MESA encodes VGG19 convolutional features as Gram matrices to capture exemplar texture, style, and stroke structure; for each neural network layer it selects the exemplar minimizing Mean-Squared Displacement (MSD) to the damaged input. Layer-wise contribution weights are derived from Optical Character Recognition-estimated character widths in the exemplar set to bias filters toward scales matching letter geometry, and a training mask preserves intact regions so synthesis is restricted to damaged areas. We also summarize prior network architectures and exemplar and single-image synthesis, inpainting, and Generative Adversarial Network (GAN) approaches, highlighting limitations that MESA addresses. Comparative experiments demonstrate the advantages of MESA. Finally, we provide a practical roadmap for choosing restoration strategies given available exemplars and metadata.
\end{abstract}



\maketitle

\input{Sections/Introduction}

\input{Sections/Related}
\input{Sections/Method}
\input{Sections/Experiments}
\input{Sections/Conclusion}

\section*{Acknowledgements}
This work is supported with image data from the research project DECAInDION (PN 1385) supported by the Hellenic Foundation for Research and Innovation (H.F.R.I.). In Memoriam: Semeli Pingiatoglou, Professor Emerita of Classical Archaeology.
\appendix
\section{Dataset Insights and Additional Experiments}
\label{sec:appendixA}

\begin{figure}[H]
    \centering
    \includegraphics[height=0.5\textheight,keepaspectratio]{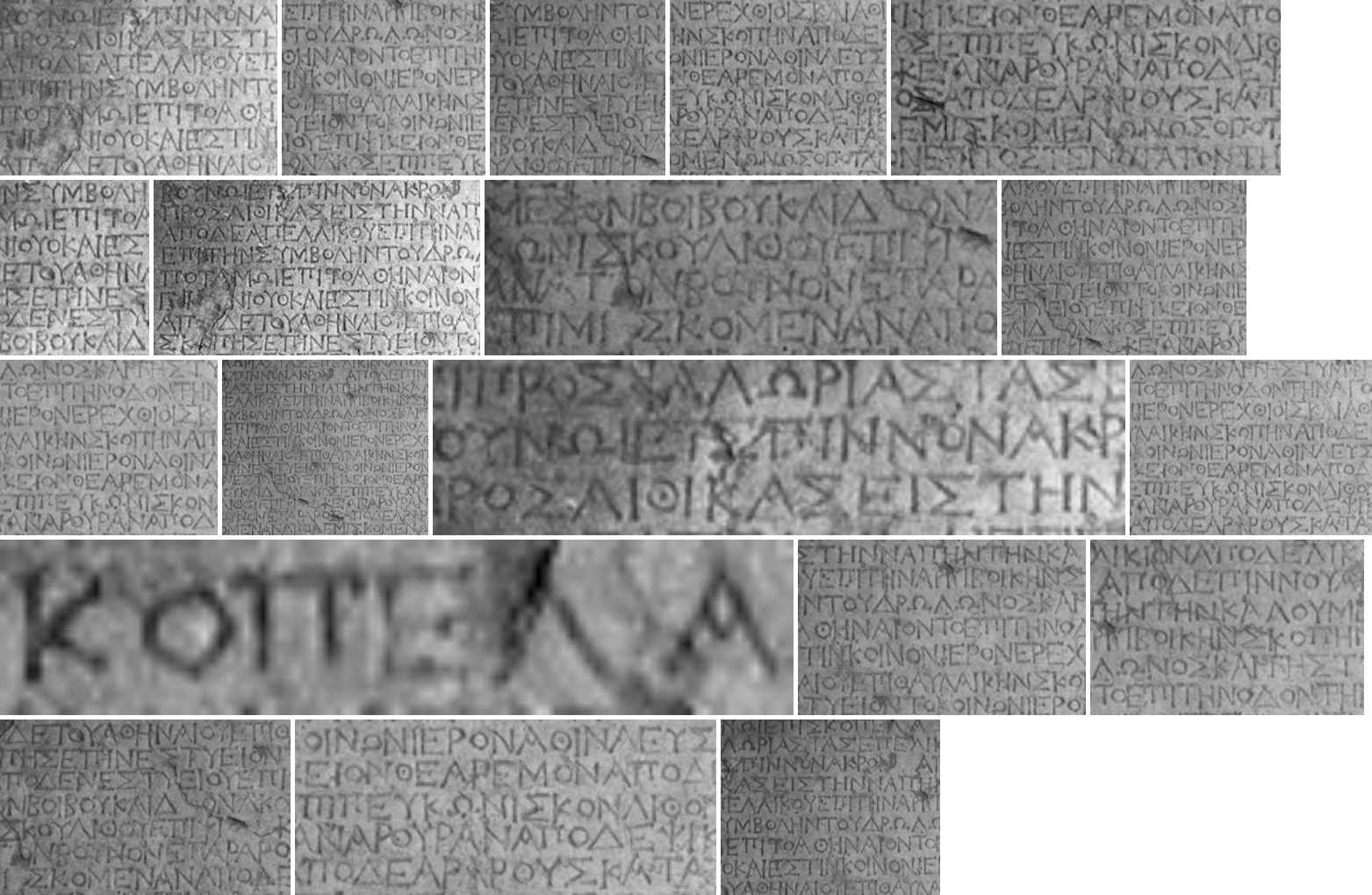}
    
    \caption{Dataset A used as exemplars in our method or as ground truth in  techniques that need supervised learning.}
    
    \Description{}
    
    \label{fig:DatasetAExemplars}
\end{figure}

\begin{figure}[H]
    \centering
    \includegraphics[height=0.48\textheight,keepaspectratio]{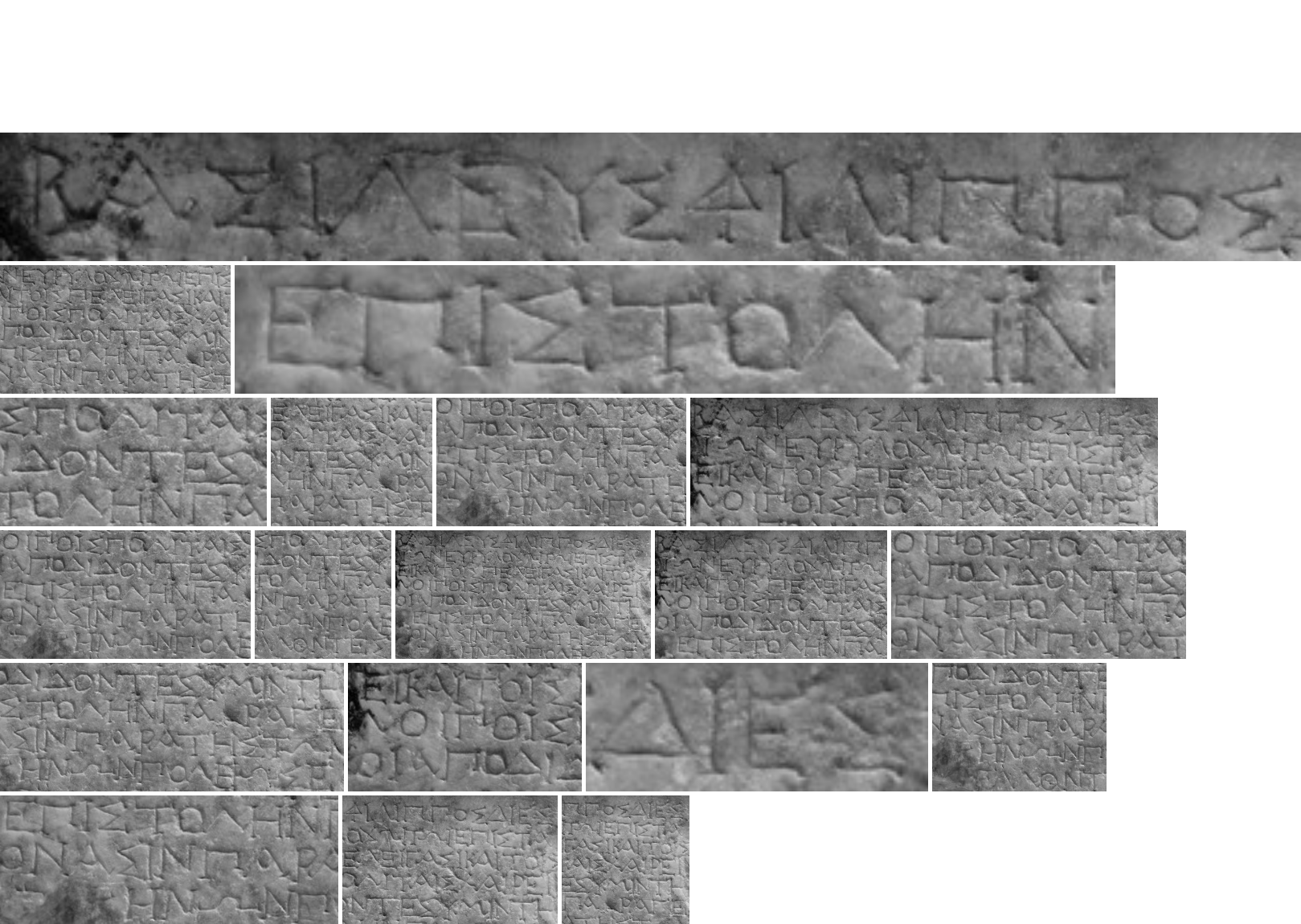}
    
    \caption{Dataset B used as exemplars in our method or as ground truth in techniques that need supervised learning.}
    
    \Description{}
    
    \label{fig:DatasetBExemplars}
\end{figure}

\begin{figure}[H]
    \centering
    \includegraphics[height=0.4\textheight,keepaspectratio]{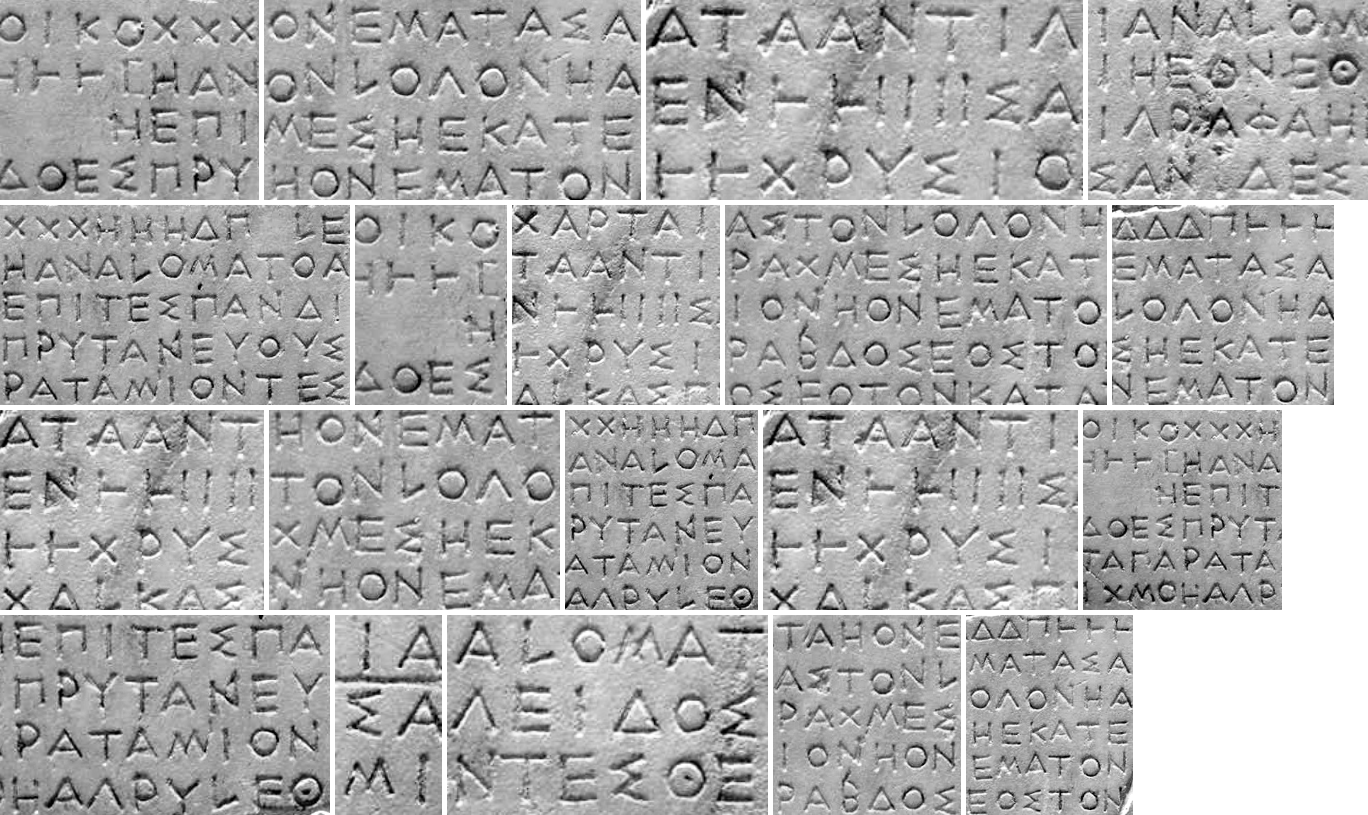}
    
    \caption{Dataset C used as exemplars in our method or as ground truth in techniques that need supervised learning.}
    
    \Description{}
    
    \label{fig:DatasetCExemplars}
\end{figure}

Figures \ref{fig:PFdatasetAResultImages}, \ref{fig:PFdatasetBResultImages} \& \ref{fig:PFdatasetCResultImages} below present our experiments trying to recover ruined inscription text by either human or environmental factors. For the methods which need training we synthetically produce training datasets. However. this is not always feasible in real world situations.

\begin{figure}[H]
    \centering
    \includegraphics[width=0.9\textwidth,height=0.83\textheight,keepaspectratio]{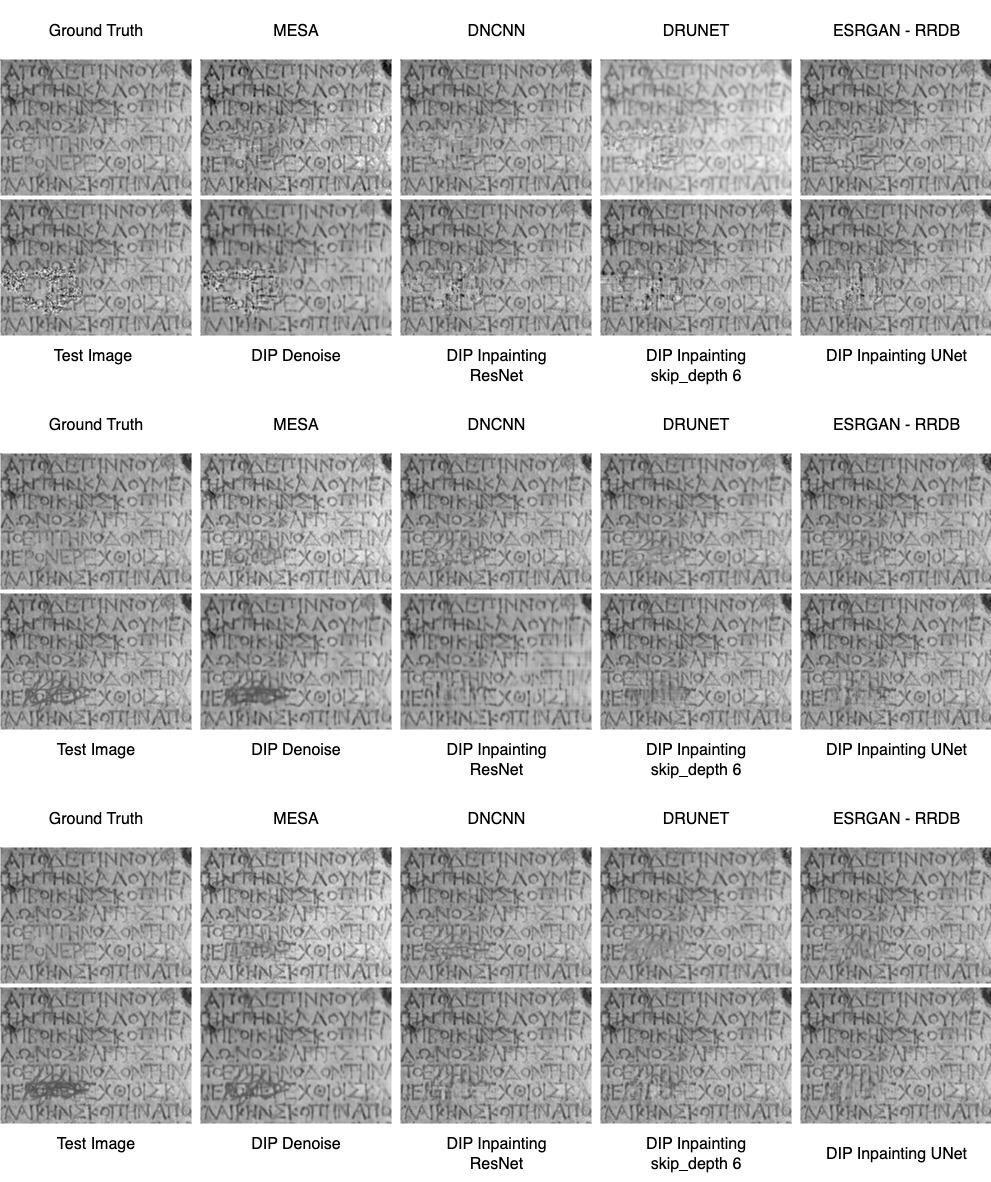}
    
    \caption{Comparison of the output of our method per test image using Dataset A images as exemplars with baseline methods.}
    
    \Description{}
    
    \label{fig:PFdatasetAResultImages}
\end{figure}

\begin{figure}[H]
    \centering
    \includegraphics[width=0.9\textwidth,height=0.83\textheight,keepaspectratio]{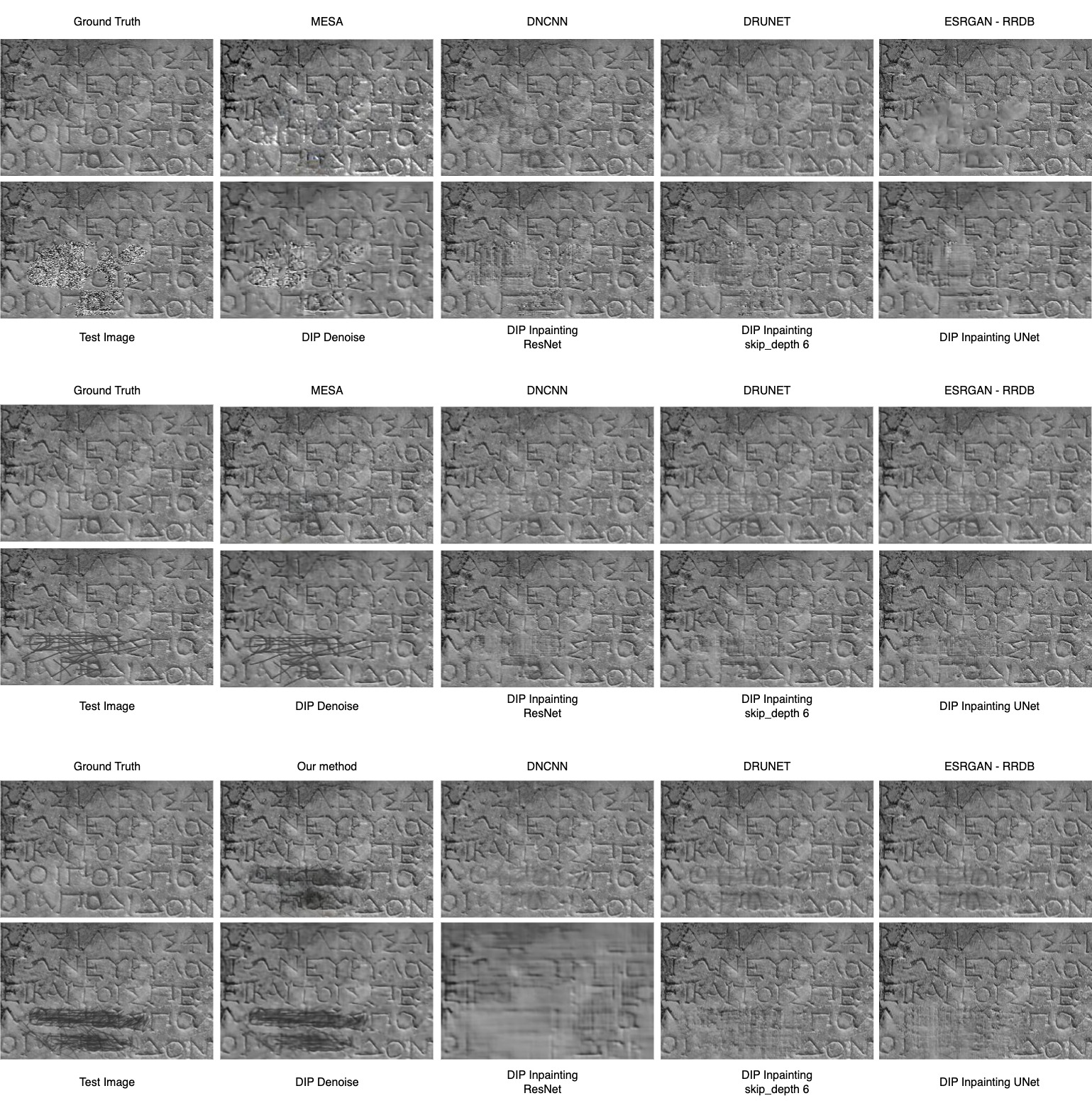}
    
    \caption{Comparison of the output of our method per test image using Dataset B images as exemplars with baseline methods.}
    
    \Description{}
    
    \label{fig:PFdatasetBResultImages}
\end{figure}

\begin{figure}[H]
    \centering
    \includegraphics[width=0.9\textwidth,height=0.83\textheight,keepaspectratio]{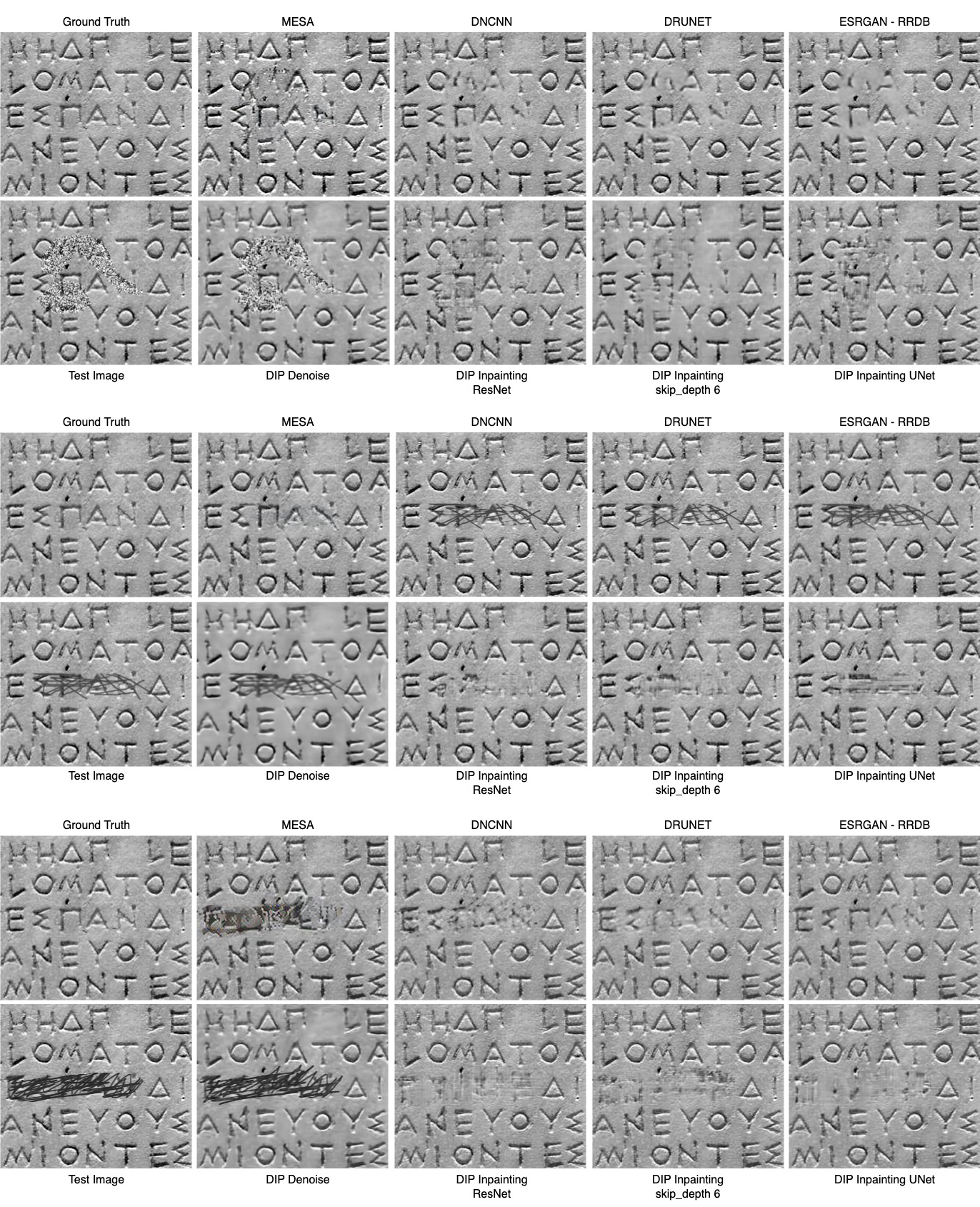}
    
    \caption{Comparison of the output of our method per test image using Dataset C images as exemplars with baseline methods.}
    
    \Description{}
    
\label{fig:PFdatasetCResultImages}
\end{figure}

\begin{figure}[H]
    \centering
    \includegraphics[page=1,width=0.9\textwidth]{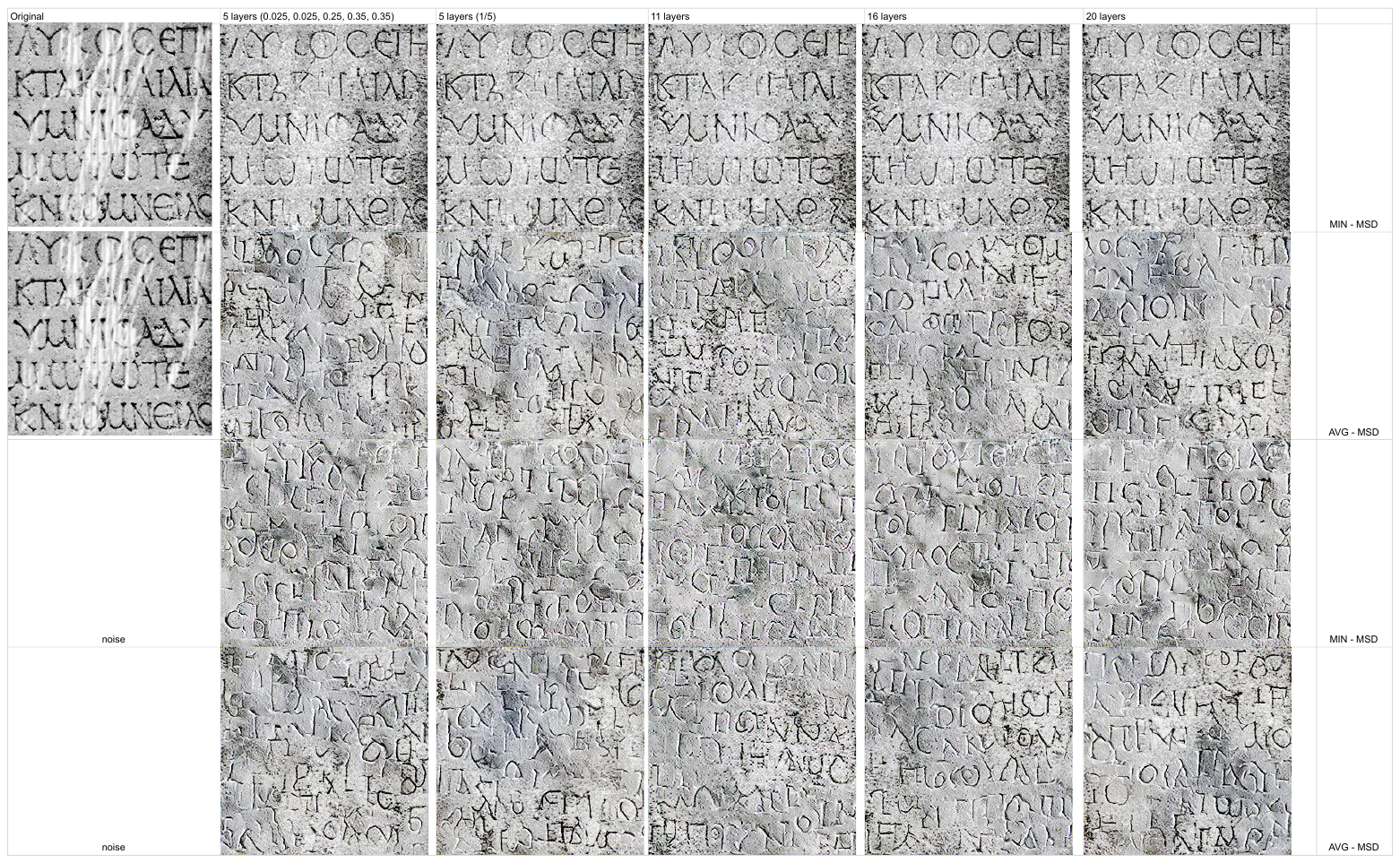}
    
    \caption{Different setups for restoring an inscription showing each network layer’s contribution and performance weight in best setup.}
    \Description{Different setups for restoring an inscription showing each network layer’s contribution and performance weight in best setup.}
    \label{fig:WhyMinFigureWith5Layers}
\end{figure}

The Figure \ref{fig:WhyMinFigureWith5Layers} shows the Input image ($1$st column), the number of layers that contribute to loss function ($2$nd-$5$th) and last columns depict the method (Average or Minimum Exemplar Gram Matrices) across all these layers used along with the loss function metric used Minimum or Average Mean Square Displacement (MSD).
In addition, results of different setups are presented where clean exemplar images of the same inscription are used to restore a ruined part of it. It appears that the best setup is using the ruined image as input with 5 contributory network layers as presented in Section \ref{sec:Method}. For completeness, this Figure additionally shows the outcomes when noise is used as input and where the optimization process guides the visual result.

\bibliographystyle{ACM-Reference-Format}
\bibliography{sample-base}

\end{document}

%% file: Sections/Introduction.tex
\section{Introduction}

Ancient inscriptions often suffer from deterioration and fading of their texture due to weathering and other environmental causes over time (erosion, fragmentation) or even due to human destruction (scratches, other types of damage). This contributes to material loss and partial or full obscuration of details, degrading the visual quality and making the text illegible.This makes the restoration of inscription images a research area domain of high interest for curators, conservators, epigraphists, historians and archaeologists. Not only are damaged inscriptions devoid of letters, but in cases the damaged or missing areas extend over multiple text segments. Additionally, some important features such as the structure and texture of inscriptions are not considered by recent methods and advances, or are only partially addressed (for example, through image-based character-level restoration \cite{mmrm}). 

In this work, we focus primarily on filling this gap by providing an image-based (material independent) technique which realizes texture enhancement while preserving the structure (size of characters, word orientation and font style) restoring only the damaged parts of it.

The technical contributions of this paper are the following:
\begin{itemize}
\item {\bf We introduce a new training free method for inscription texture (image) restoration.}
We employ a VGG19 \cite{vgg19} neural network on a multi-exemplar training schema for recovering missing or damaged parts of inscription textures. The multi-exemplar learning process helps on restoration leveraging each contributing neural network layer via extracting a feature map with the usage of Gram Matrices and computing a minimum style loss (normalized Mean Square Error) of them both for the input and the exemplar images. In this manner, we guide the learning process on finding the closest exemplar patterns to restore the damaged or missing inscription texture segments. This approach is feasible because of an observation in such methods (see e.g. \cite{deepTerrainExpansion}, \cite{deepTiling}) that suggests that there is no limitation on using exemplars of different sizes. 
Thus, using exemplars of not equal dimensions we have developed a framework that matches the best exemplar candidates on each layer and training step guiding the learning optimization process to a rebuilt of its missing parts inscription texture.
Moreover, we propose a new way of calculating the weights for each layer contribution during the learning process. The new novelty is that the calculation is based on matching filter size with the letters width. A distribution of letter widths is utilized using Optical Character Recognition (OCR) and more specifically Tesseract \cite{tesseract} so that filter sizes in neural network layers match distribution probabilities and weights are extracted from the distribution itself. Thus, instead of arbitrary weights assignment, our approach matches the filter sizes of the network with the distribution of letter widths of both the damaged input and exemplars images. It extracts probabilities by fitting a distribution to the letter widths that form the layer weights. In this way, we ensure that the receptive fields of the neural network are guided by the structural characteristics of the inscriptions.

 \item{\bf We adapt and evaluate previous image enhancement techniques for inscription texture (image) restoration}. We have adapted several image enhancement techniques for inscription image restoration. We examine the requirements for training with or without priors and we evaluate the effectiveness of each method on inscriptions. 

\item{\bf We introduce new metrics for evaluating inscription texture (image) enhancement.}
This paper additionally contributes to text recovery metrics by proposing two new metrics. One is a log-scaled version of Levenshtein Distance \cite{levenshtein} as a similarity metric that strictly limits the value to $(0,1]$ and another alternative using a normalized Levenshtein Distance version using a fixed denominator defined as the worst case operations scenario (in our work $100$) to limit which are the most operations needed for a recovered text to reach the original (ground-truth) text. In the latter metric case, if it exceeds $1$ we cap it to $1$. Equivalently, any recovery distance greater than $100$ is effectively treated as $100$.
\end{itemize}

Section 2 provides a short overview of related work. Section 3 provides an overview of methods for image inscription restoration and introduces our {MESA-TF} method. Section 4 provides an experimental comparative evaluation of all inscription image restoration methods. Finally, Section 5 offers conclusions and future research directions.

%% file: Sections/Related.tex
\section{Related Work}

Recent methods in the field of text inscription restoration have primarily focused on utilizing large language models (LLMs) or have been limited in scope, often emphasizing character-level reconstruction. Notable character-level methods include Pythia \cite{pythia}, while transformer-based models such as Ithaca \cite{ithaca} and Aeneas \cite{aenas} attempt to predict missing characters from the surrounding context. Fine-tuned Llama 3.1 \cite{llama}, an instruction-tuned LLM, also focuses on predicting missing or illegible characters by leveraging grammatical and contextual patterns. Multimodal and multitask models like MMRM \cite{mmrm}, despite leveraging visual features, primarily operate at the character level, combining contextual information with visuals but are largely aimed at the Chinese language. These methods exhibit efficient character reconstruction but encounter significant drawbacks at the sentence-image level, lacking the ability to recover both text and visual structures of inscriptions in a manner that preserves the appearance of damaged inscriptions. Our method addresses this by restoring inscriptions at the image level through deep exemplar-based learning and a masking strategy, effectively restoring damaged regions while preserving intact areas.

In parallel, image restoration, style transfer, denoising, and inpainting frameworks have significantly advanced in recent years. In this section, we discuss well-known frameworks and highlight their contributions and relevance to our work.

Texture synthesis has been a highly researched area over the last decade, with fields such as super-resolution, expansion, and restoration drawing significant attention. A pivotal work in texture synthesis is by Gatys et al. \cite{gatys}, which utilized neural style transfer and leveraged pre-trained Convolutional Neural Networks (CNNs such as VGG16, VGG19) for feature extraction and comparison. By minimizing a style loss using Gram matrices of features extracted from deep Convolutional Neural Networks, this approach preserves semantic structure while generating visually acceptable textures, a key contribution also applicable in restoration. Our framework adopts perceptual loss to guide the enhancement of degraded parts of inscriptions, utilizing exemplars of varying sizes for high accuracy and efficiency.

Recent advances in texture synthesis focus on different aspects and achieve high visual standards. State-of-the-art methods include ESRGAN \cite{esrgan}, DNCNN, DRUNET, FFDNet, and SwinIR. ESRGAN, an extension of SRGAN \cite{srgan}, introduces the Residual-in-Residual Dense Block (RRDB) to improve training stability and reduce artifacts, combining perceptual and adversarial losses computed from deep features to generate high-fidelity textures. DNCNN (Denoising Convolutional Neural Network) \cite{DNCNN} targets image denoising by predicting residual noise, while DRUNET \cite{drunet}, an U-Net architecture with residual blocks and internal connections, serves as a general-purpose restoration network. These networks, along with FFDNet \cite{ffdnet} and SwinIR \cite{swinr}, showcase the power of deep learning in feature extraction for robust image restoration. In our pipeline, we utilize RRDB-based generators, DNCNN, and DRUNET as mechanisms for inscription restoration, training them on damaged and ground-truth inscriptions. Other methods for universal style transfer methods such as Whitening and Coloring Transform (WCT) \cite{WCT} are effective  for visual style synthesis. However, their direct application to inscription restoration is limited because of degradation patterns being not able to meaningfully treated as transferable styles. Exemplar-conditioned style transfer methods, such as AdaIN-based \cite{ADAIN} approaches, introduced the idea of using reference images to guide feature statistics. Their focus on artistic style makes them only marginally relevant to MESA, as the concept of exemplar-based guidance for controlling local appearance and consistency is applicable, but the artistic style transfer aspect creates a conceptual gap.

Denoising techniques play a fundamental role in improving the quality of ancient inscriptions, which often contain noisy segments. DNCNN \cite{DNCNN} effectively reduces noise while preserving undamaged parts, improving the readability of damaged text. Image inpainting, which fills in missing or unwanted fragments, has progressed from patch-based synthesis to deep architectures and generative priors. U-Net \cite{unet} encoder-decoder models with skip connections excel due to their efficiency in preserving structural details. Partial Convolution Inpainting \cite{pconv} enhances convolutional layers to handle arbitrary textures, while Deep Image Prior (DIP) \cite{deepImagePrior} reconstructs selected areas without large datasets, relying on the network itself as a prior. Our framework evaluates U-Net, ResNet, and skip-connection inpainting models using DIP, specifically adapted for inscription restoration. LaMa (Large Mask inpainting with Fourier convolutions) \cite{LaMa}, another state-of-the-art method, uses fast Fourier convolutions for large receptive fields but is unsuitable for our inscription restoration due to its dependency on synthetic missing regions rather than naturally damaged images. LaMa demands synthetic masks to be trained, learning to fill in artificially removed regions. Consequently, training LaMa becomes prohibitively expensive too.

Other approaches of diffusion-based document restoration such as DiffHDR \cite{DiffHDR} or synthetic paired-data pipelines such as PreP-OCR \cite{PreP-OCR} have demonstrated strong results in restoring degraded documents. These methods are focusing on character-level fidelity and they propose useful evaluation practices, such as measuring accuracy in repaired regions. Both MESA and the aforementioned methods aim to improve fidelity in degraded regions but they differ fundamentally in structural fidelity. DiffHDR and PreP-OCR do not leverage exemplar images and may distort or smooth out small structural details contrary to MESA which uses exemplar guidance to restore local structures. This makes MESA more accurate on restoration of fine-grained patterns such as letters in inscriptions.

%% file: Sections/Method.tex
\section{Deep Inscription Texture Restoration}
\label{sec:Method}

In this work, we deploy well-known methods on image inscription restoration with the usage of degraded images (with deteriorated texture) and original images (as ground-truth) and we propose a novel deep-learning method based on VGG19 \cite{vgg19} within a multi-exemplar framework to generate missing parts of damaged inscription texture parts and thereby recover lost letters and content. All of these approaches are material-independent (they do not require prior knowledge of the physical properties or composition of the objects in the image such as clay, metal, glass, bone, mosaic). They operate entirely at the image level via extracting and enhancing features.

\subsection{Deploying Image Enhancement Methods for Inscription Restoration}

In our work, we deploy ESRGAN \cite{esrgan}, DRUNET \cite{drunet}, and DNCNN \cite{DNCNN} as supervised restoration networks for texture recovery of damaged inscriptions. Each network is trained on degraded and ground-truth image pairs where artificially introduced damage are applied to clean inscription textures to simulate damaged inscription fragments.

\begin{figure}[H]
    \centering
    \includegraphics[width=0.9\textwidth,keepaspectratio]{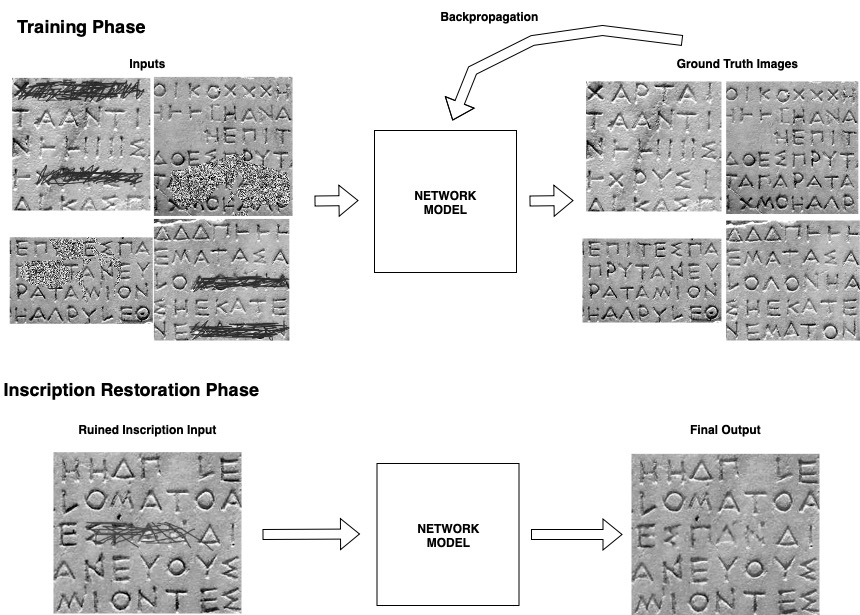}
    
    \caption{Training and Feed-Forward Phases for Inscription Restoration of well-known Network Models (DRUNET, ESRGAN Generator part and DNCNN)}
    
    \Description{Training and Feed Forward Phases for Inscription Restoration of well-known Network Models (DRUNET, ESRGAN Generator part and DNCNN)}
    
    \label{fig:wellknownmethodstrained}
\end{figure}

Ground-truth images provide supervision for reconstructing degraded parts of damaged inscriptions. DNCNN focuses on noise reduction which sets it as a better candidate for reconstructing noisy patches while DRUNET focuses on structure preservation making it more effective on human-caused damage to inscriptions (e.g scratch removals). ESRGAN RRDB-based generator enables inscription texture restoration utilizing perceptual only losses variations for improving ground-truth "look and feel". The restoration process and how such networks are deployed in our inscription pipeline is presented in Figure \ref{fig:wellknownmethodstrained}. In all network cases, the training phase is conducted with damaged and ground-truth paired inscription textures (supervised learning).

In contrast, Deep Image Prior (DIP) \cite{deepImagePrior} is employed as a training-free baseline. A randomly initialized Convolutional Neural Network (CNN) is optimized independently for each corrupted input image, using only the observable pixels as a self-supervised signal to reconstruct the missing inscription regions. In this framework, the prior knowledge about clean, structured content and smooth intensity transitions is implicitly encoded by the network architecture itself. This architectural prior guides the optimization process, enabling a noise-like input image to be mapped to an output that preserves the intact regions while attempting to reconstruct the degraded inscription areas or fragments.

\subsection{MESA-TF: A novel Multi‑Exemplar, Style‑Aware, Training Free Method for Inscription Texture Restoration}

In style transfer or terrain generation, during the training phase with backpropagation, the input images can be of arbitrary spatial resolution when distances between Gram matrices are employed as the style similarity measure and these matrices are computed from the feature maps of a VGG19 network \cite{vgg19}. This is possible because the dimensionality of a Gram matrix depends solely on the number of filters in each network layer, rather than on the spatial dimensions of the input, an observation initially reported in \cite{deepTerrainExpansion}. In our method, we adopt and further extend this concept by introducing a multi-exemplar framework that correlates feature representations across layers to synthesize images exhibiting similar style characteristics. These correlations are computed from the feature maps following the formulation in \cite{gatys}, and the optimization is carried out using the L-BFGS (Limited-memory BFGS) algorithm \cite{LBFGS}. 

The feature representations between the input image and the exemplar images at a given network layer are based on feature map activations denoted by $F^l_{ab}$, where $F^l_{ab}$ is the activation of the $a$-th filter at spatial position $b$ in layer $l$. As described in \cite{gatys}, this formulation enables the model to capture patterns where individual spatial positions are interpreted as instances of a more general feature representation. Formally, the feature maps at layer $l$ can be arranged into a matrix $F^l \in \mathbb{R}^{N^l \times K^l}$, where $N^l$ is the number of filters and $K^l$ is the flattened spatial dimensionality of each feature map. However, in our setting we are not primarily interested in exact spatial correspondences, but rather in style similarity so that damaged texture regions can be plausibly restored. Consequently, raw feature maps are insufficient, and Gram matrices must be employed to capture spatially invariant second-order statistics of the features. A Gram matrix is defined as:

\begin{equation}
G^l_{ab} = \sum_i F^l_{ai} F^l_{bi}
\end{equation}

\begin{figure}[H]
    \centering
    \includegraphics[width=0.9\textwidth,height=0.83\textheight,keepaspectratio]{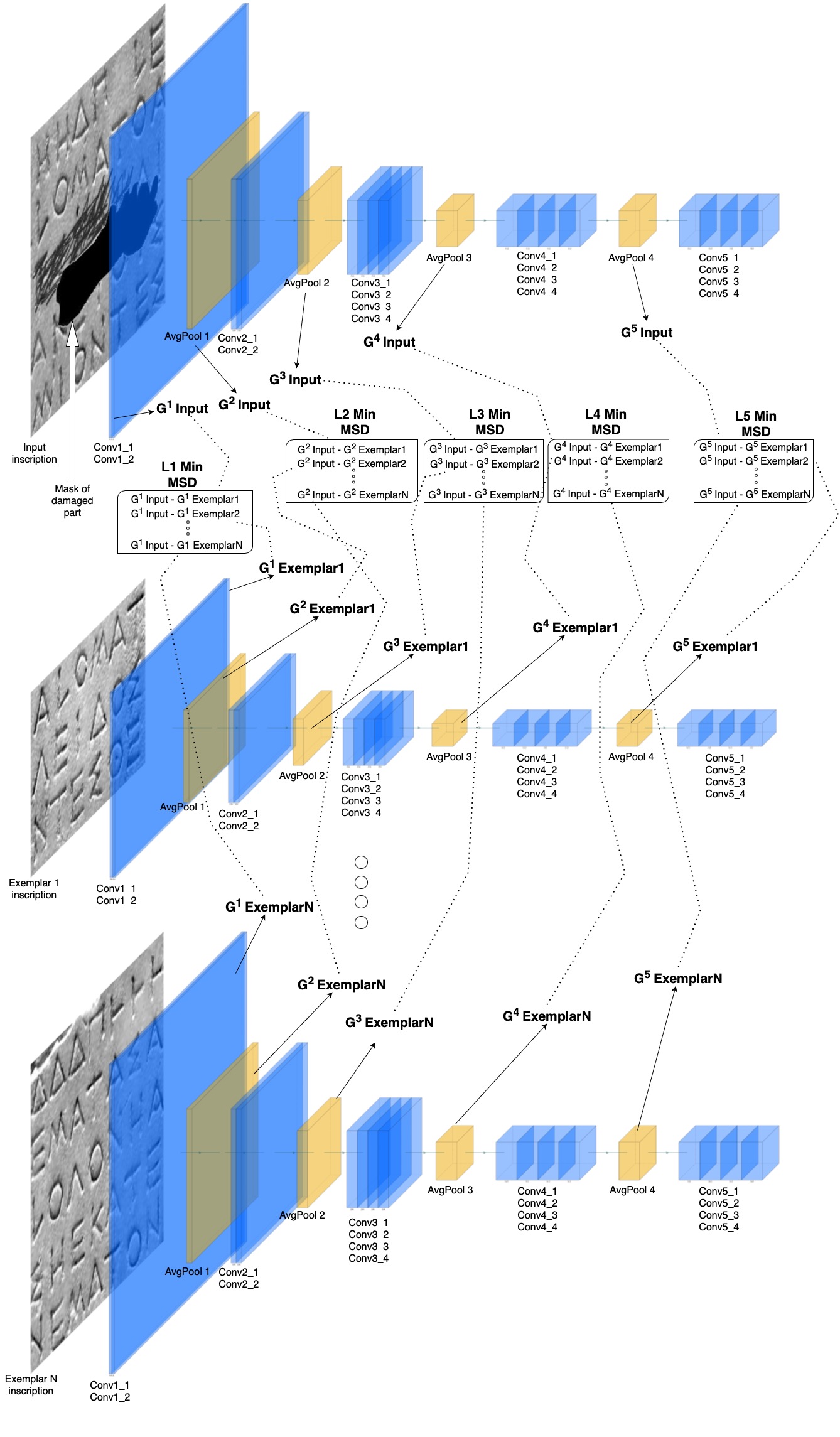}
    
    \caption{Restoring Ancient Inscriptions: $G$ denotes the Gram matrix of feature maps in layer $l$, depending on the number of filters $F_l$. The network structure is based on VGG19 \cite{vgg19}, using AvgPooling instead of MaxPooling layers. Networks generated with PlotNeuralNet (https://github.com/HarisIqbal88/PlotNeuralNet) and then modified.}
    
    \Description{Restoring Ancient Inscriptions. $G$ denotes the Gram matrix of feature maps in layer $l$, depending on the number of filters $F_l$. The network structure is based on VGG19 \cite{vgg19}, using AvgPooling instead of MaxPooling layers. Instances in the figure were generated with PlotNeuralNet (https://github.com/HarisIqbal88/PlotNeuralNet) and then modified.}
    
    \label{fig:network}
\end{figure}

As a consequence, for each exemplar the corresponding part of the loss function is defined as:

%
\begin{equation} L_{exemplar_i}(I_{damaged}, I_{exemplar_i}) =   \frac{1}{4 (N^l)^2 \, (K^l)^2}  \sum \big(G^l_{damaged} - G^l_{exemplar_i}\big)^2 \end{equation}

where $I_{damaged}$ is the inscription texture to be recovered and $I_{exemplar_i}$ 
is one of the exemplar textures. On each layer we keep the minimum loss value $L_{l}^{\text{Min}}$ where $l$ symbolizes the $l$-th contributory layer along all pairs of degraded inscription and training exemplars (in Figure \ref{fig:network} each layer's final loss value is noted as the sum minimized over all candidates $L_{\text{exemplar}_i}$) and Layer loss is calculated as:

\begin{equation}
L_{l}^{\text{Min}}
= \min_{i}^{E} \; L_{\text{exemplar}_i}(I_{\text{damaged}}, I_{\text{exemplar}_i}).
\end{equation}

where $E$ is the number of exemplars for which we calculate all $L_{\text{exemplar}_i}$. Then the total loss function is formulated as:

\begin{equation}
L_{total}
= \sum_{l}^{N_{all}} \; w_l L_{l}^{\text{Min}}.
\end{equation}

where $N_{\text{all}}$ is the number of layers included in the loss and $w_l$ is the weight associated with layer $l$.


Our method correlates feature maps from layers (see Figure \ref{fig:network}) $layer1, AvgPool1, AvgPool2, AvgPool3$, and $AvgPool4$. We report experiments in Appendix \ref{sec:appendixA} (see in particular Figure \ref{fig:WhyMinFigureWith5Layers}, which shows that, for inscription restoration, these layers contribute more to producing visually high-quality results. Each layer's contribution to the loss is weighted by a corresponding factor; the procedure used to determine these weights is described in Subsection~\ref{sec:layerContribution}.


In Figure\ref{fig:network} the MESA architecture is presented. For each layer contributing to the loss function, we compute the Gram matrices of both the input inscription texture and all training exemplars, using structurally identical network instances. In a second step, we evaluate the mean squared distance between the Gram matrices of the input and each exemplar, and we retain, for each layer, the minimum distance as that layer’s contribution to the main style loss. In this way, we encourage similar feature representations across the selected contributing network layers. The damaged image is fed to the first network instance in Figure \ref{fig:network}, while the remaining instances receive the clean training exemplars, thereby forcing the network to minimize the discrepancy between the exemplars and the input image by replacing degraded texture regions.

Masking the image, both at initialization and at each training step in either of the two training procedures, plays a crucial role in preserving color and structural information during the learning phase. In Figure \ref{fig:network}, the mask is shown on the initial input, which is particularly important for maintaining contrast, luminance and the structure of intact inscription regions, while directing the framework to focus primarily on restoring the damaged or degraded parts of the texture. Moreover, for some inputs, applying the same mask at every iteration leads to more faithful reconstructions of the inscriptions.

More specifically, a spatial masking strategy is applied at the image level to restrict the optimization to the damaged region while preserving the intact content. At each optimization step, the current estimate is composited with the initial image using a binary mask such that only pixels inside the masked region are allowed to change. The pixels outside the mask are clamped to their original values at every training step as well. The resulting composite image is then forwarded through the network for feature extraction and by using Gram matrices, exemplar-based losses are calculated globally over the entire spatial extent of each layer. Although the Gram matrices capture global feature correlations, the optimization affects exclusively the masked region owing to the fact that gradients are effectively suppressed by the image-level masking with respect to the unmasked pixels. This design enforces texture and style consistency with the best-fitted exemplar while ensuring that modifications are merely restricted to the ruined area.

\subsection{Layer contribution - Character-Scale Weighting}
\label{sec:layerContribution}

\begin{figure}[htbp]
    \centering
    
    \begin{subfigure}[b]{0.4\textwidth}
        \centering
        \includegraphics[height=3.2cm]{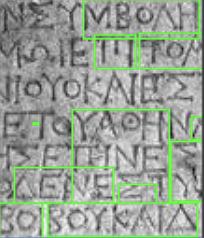}
        \caption{Dataset A}
        \label{fig:aeaDataset}
    \end{subfigure}
    \hfill
    \begin{subfigure}[b]{0.5\textwidth}
        \centering
        \includegraphics[height=3.2cm]{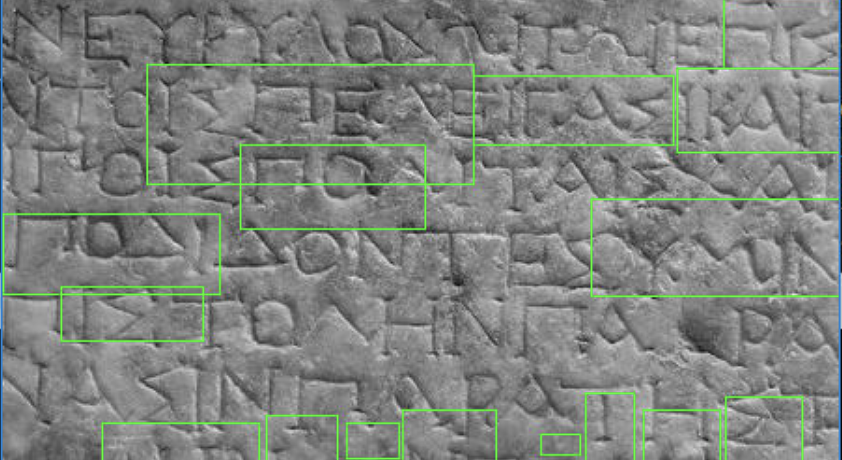}
        \caption{Dataset B}
        \label{fig:arxaiaellinikaDataset}
    \end{subfigure}
    
    \vspace{1em}
    
    \begin{subfigure}[b]{0.4\textwidth}
        \centering
        \includegraphics[height=3.2cm]{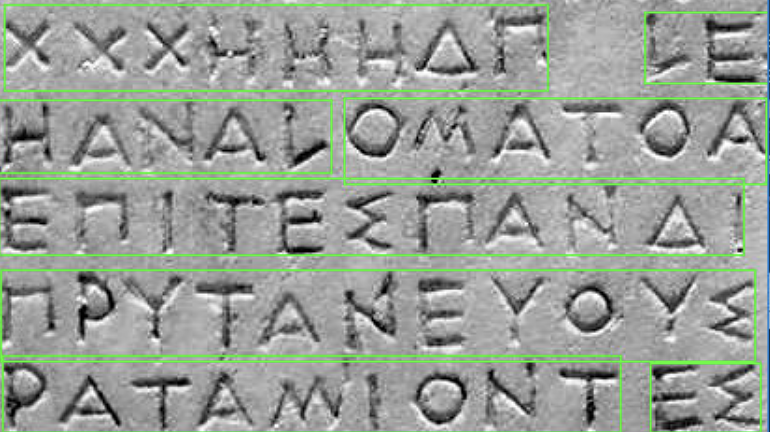}
        \caption{Dataset C}
        \label{fig:aeDataset}
    \end{subfigure}
    
    \caption{Tesseract \cite{tesseract} letter widths detection. Each rectangle captures some letters and we divide each width by the number of letters detected within it to define an average width of letters inside it.}
    \Description{Comparison of fitted distributions across datasets.}
    \label{fig:tesseractLetterWidths}
\end{figure}

Moreover, our work introduces an innovative, data‑driven weighting scheme for each contributing network layer in the loss function, instead of relying on fixed, hand‑tuned weights. We experiment with different values per layer, as well as with their combinations. An OCR‑based technique is employed to perform a statistical analysis of the letter widths in both the ruined input inscription image and the clean exemplar inscription textures, as illustrated in Figure \ref{fig:tesseractLetterWidths}. We use Tesseract \cite{tesseract}, which has been further trained on our specific datasets (consisting of the input images and their corresponding transcriptions) as presented in Section \ref{sec:appendixA}. This additional training improves text recognition on our datasets and as a consequence enables the ability for more accurate weight calculations based on the widths of the letters. Our letter width analysis, performed via OCR, is not prone to the typical inaccuracies of full text recognition. This is because (i) the OCR process is applied only to clear, well-preserved exemplars of the inscription. (ii) Furthermore, our statistical approach focuses on extracting robust geometric features (like bounding box widths) rather than relying on perfect character identification. This aggregation of geometric data is inherently more resilient to noise and minor imperfections that would otherwise compromise overall OCR performance.


 
\begin{figure}[htbp]
    \centering
    
    \begin{subfigure}[b]{0.45\textwidth}
        \centering
        \includegraphics[height=5cm]{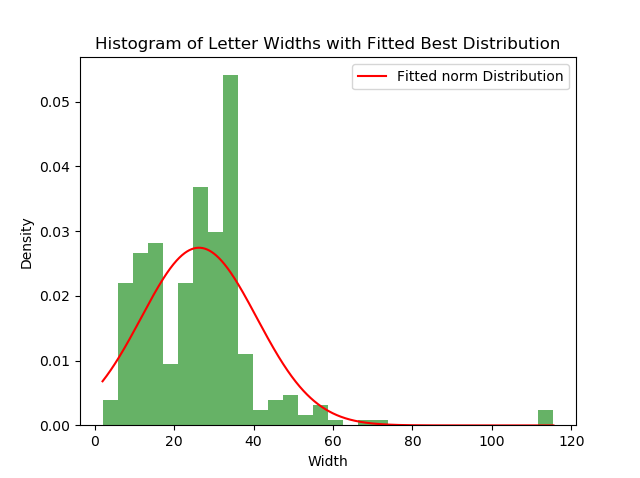}
        \caption{Dataset A}
        \label{fig:ae}
    \end{subfigure}
    \hfill
    \begin{subfigure}[b]{0.45\textwidth}
        \centering
        \includegraphics[height=5cm]{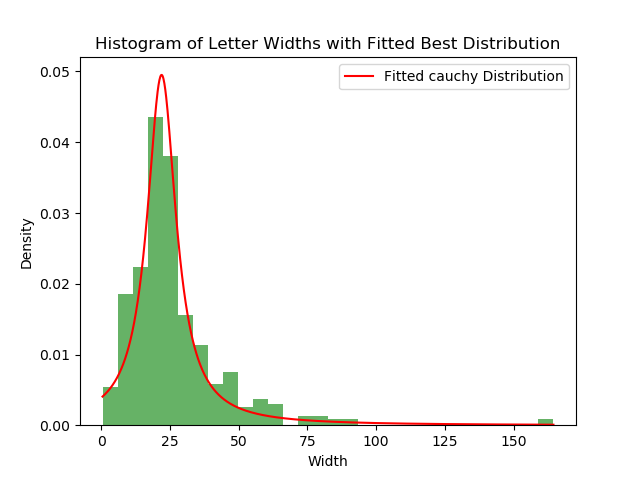}
        \caption{Dataset B}
        \label{fig:aea}
    \end{subfigure}
    
    \vspace{1em} 
    
    \begin{subfigure}[b]{0.6\textwidth} 
        \centering
        \includegraphics[height=5cm]{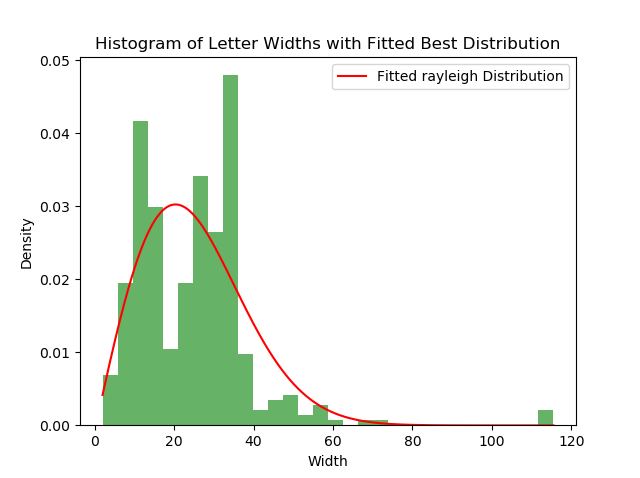}
        \caption{Dataset C}
        \label{fig:arxaiaellinika}
    \end{subfigure}
    
    \caption{Best-fit distributions of letter widths for the three datasets used in our experiments.}
    \Description{}
    \label{fig:differentDatasetDistributions}
\end{figure}

The contribution of each network layer to the total loss is modulated by the empirical distribution of letter widths in our dataset. 
Let $P(w)$ denote the probability density function (PDF) of letter widths $w$, obtained from the best-fitting distribution over both the input and training exemplars (see Figure~\ref{fig:differentDatasetDistributions}). 
For a given layer $l$ with filter (or width) size $w_l$, we define its contribution weight, $weight_l$, as the cumulative probability of letter widths falling within a distribution interval determined by the mean $\mu$ and standard deviation $\sigma$ (e.g.\ $[\mu, \mu + \sigma], [\mu, \mu + 2\sigma], [\mu - \sigma, \mu]$).

Let \(F(w)\) be the cumulative distribution function (CDF) of letter widths, with mean \(\mu\) and standard deviation \(\sigma\).  
We partition the distribution into intervals defined by multiples of the standard deviation around the mean.
For a given interval \(W_l = [a_l, b_l]\), the probability of a letter width falling inside it is
\[
P(w \in W_l) \;=\; F(b_l) - F(a_l).
\]

This represents the probability that the letter widths lie within the layer's width intervals. Consequently, this probability is used as the weight (contributory factor) of the layer $l$. To ensure that the contribution of each layer is normalized, we compute normalized weights $\hat{{weight}_l}$ as

\[
\hat{{weight}_l} = \frac{{weight}_l}{\sum_{k=1}^{L} {weight}_k},
\]

where $L$ is the number of layers that contribute to the loss. The final total loss is then computed as a weighted sum defined as

\[
\mathcal{L}_{\text{total}} = \sum_{l=1}^{L} \hat{{weight}_l} \, L_{l}^{\text{Min}}.
\]

where $L_{l}^{\text{Min}}$ is the layer's min exemplar loss as described above.

Thus, we employ a novel, data-driven, statistically informed weighting scheme that ensures that the loss contribution of each layer is proportional to how well its filter width aligns with the statistical distribution of letter widths in our dataset.

\subsection{A Roadmap for Inscription Texture Restoration}

Each method is shown to perform better in specific scenarios. 



MESA is a training-free method for inscription restoration, relying solely on pre-trained generic deep learning feature extractors (e.g., from a VGG19 network) rather than requiring new training on inscription-specific data. MESA is particularly suitable when the damaged inscription is part of a larger text containing intact regions that can serve as exemplars, or when a dataset of inscriptions in the same or a similar style is available. In contrast, Deep Image Prior is more appropriate when only a single inscription image is available and its intact regions are insufficient to provide effective data for our multi-exemplar strategy.

By comparison, DRUNET, DnCNN, and ESRGAN require either (i) a real-world dataset with images before and after damage—which is rarely available for ancient inscriptions—or (ii) a synthetic dataset capable of accurately reproducing, on other well-preserved inscriptions, the type of damage present in the degraded input. Moreover, each of these methods is better suited to specific types of degradation. DRUNET is more effective for human-induced damage such as scratches, which are common on ancient inscriptions, whereas DnCNN is best suited for reducing noise caused by environmental factors. ESRGAN, originally designed for image super-resolution, is preferable when high-resolution restoration is required in ruined regions. Nonetheless, our experiments show that ESRGAN performs comparably well even when the output resolution is constrained to match that of the input image.

Table \ref{tab:methods-comparison} summarizes the training strategy for each method along with a brief description.

\begin{table}[H]
\centering

\resizebox{0.95\linewidth}{!}{%
\begin{tabular}{|l|c|p{5cm}|}
\hline
\textbf{Method} & \textbf{Training Strategy} & \textbf{Description} \\
\hline
\textbf{MESA} &  Multi-exemplar, Per-input Optimization, Training Free & Multi-exemplar style-aware method without pretraining; ideal for inscription restoration when paired data are unavailable (e.g., denoising, inpainting). \\ 

\textbf{Deep Image Prior (DIP)} & Self-supervised (single image), Per-input optimization, Training Free & Learns directly from a single corrupted image; effective for denoising, inpainting when dataset is not available. \\ 

\textbf{DRUNET} & Supervised (pretrained) & Requires dataset pretraining; robust for general restoration (denoising, deblurring, artifact removal). \\ 

\textbf{DNCNN} & Supervised (pretrained) & CNN trained on large datasets; is highly effective for Gaussian denoising and overall noise removal. \\ 

\textbf{ESRGAN (generator part)} & Supervised (pretrained) & Trained on damaged/ground truth paired inscription images; best for super-resolution and perceptual enhancement while in this work we use only the generator part of it. \\ 
\hline
\end{tabular}%
}
\caption{Comparison of image restoration methods regarding training strategy and suitable use cases.}
\label{tab:methods-comparison}
\end{table}

%% file: Sections/Experiments.tex
\section{Experiments \& Results}

We conducted experiments on three different inscription datasets. Each dataset is divided into scratched and noise subsets. These datasets are used in our method both as clean exemplars and as ground-truth images for supervised baseline methods, which require paired ruined and clean images for training (see Figures~\ref{fig:DatasetAExemplars}, \ref{fig:DatasetBExemplars}, and \ref{fig:DatasetCExemplars}). In addition, synthetic datasets of ruined inscriptions, affected by either noise or scratches (see the aforementioned figures with ruined segments), are generated for the training phase of these supervised methods, ensuring a fair comparison.

\subsection{Metrics}
\label{subsec:SubSectionMetrics}

Some well-known metrics that we used to evaluate our method are PSNR\cite{PSNR}, SSIM\cite{SSIM}, and LPIPS\cite{LPIPS}. Peak Signal-to-Noise Ratio (PSNR) measures how close the pixels of the restored inscription are to those of the ground-truth image. Moreover, the Structural Similarity Index (SSIM) between two images captures perceptual aspects such as luminance, contrast, and structural information.
On the other hand, a metric that is closer to human visual judgment is the Learned Perceptual Image Patch Similarity (LPIPS), a deep learning-based perceptual measure that uses correlations of deep neural network features to compute a similarity score.

Another metric that we employ, both for evaluating our method and for making the newly proposed metrics presented below more interpretable and their assessment more comprehensive, is the Levenshtein distance \cite{levenshtein}. This metric measures the minimum number of character-level operations required to transform the recovered text into the original text.

Our work additionally contributes to the field of text restoration by introducing two new metrics which can be combined with image quality metrics bounded to $[0,1]$ (e.g. SSIM, PSNR, etc).
The first is called Text Recovery Score (TRS) which uses a Normalized Edit Distance which fixes the normalization denominator at $100$ edits. To prevent values greater than $1$ we cap the score at $1$. This produces a bounded metric in the interval $[0,1]$ ensuring that very long or highly corrupted texts do not dominate the evaluation. Formally:

\begin{equation}
\text{TextRecoveryScore}(o,r) = 1 - \min\left(1, \frac{LD(o,r)}{S}\right)
\end{equation}
\noindent
where $LD(o,r)$ is the Levenshtein distance \cite{levenshtein} between the original text $o$ and the recovered text $r$, and $S$ is a fixed positive integer that specifies the maximum number of edit operations considered. In our experiments, we set $S = 100$, which is a reasonable choice in many cases, clamping all larger distances to $100$. A perfect match yields $0$, while any distance $\geq 100$ is effectively treated as $100$. The resulting values are bounded in $[0,1]$.

To illustrate the advantage of TRS over the raw Levenshtein Distance, consider the example shown in Table~\ref{tab:trs_comparison}. $In the first case, a long textual sentence contains a moderate number of OCR-induced substitution errors, resulting in an edit distance of $18$. In the second case, the same sentence is paired with a nearly random character sequence, corresponding to an extreme recognition failure with an edit distance exceeding 100.$ While the raw Levenshtein Distance grows unbounded and disproportionately penalizes such catastrophic failures, TRS assigns a moderate score to the partially recovered text and clamps the completely corrupted output to a score of $0$. This behavior ensures that realistic variations in OCR quality are preserved, while preventing extreme outliers from dominating aggregated evaluation results.

\begin{table}[H]
\centering
\setlength{\tabcolsep}{8pt}
\caption{Comparison between Levenshtein Distance (LD) and the proposed Text Recovery Score (TRS) with fixed normalization ($S=100$).}
\label{tab:trs_comparison}
\begin{tabular}{>{\raggedright\arraybackslash}p{3.5cm} >{\raggedright\arraybackslash}p{3.5cm} c c >{\raggedright\arraybackslash}p{4.6cm}}
\toprule
\textbf{Original Text} &
\textbf{Recovered Text} &
\textbf{LD} &
\textbf{TRS} &
\textbf{Error Analysis} \\
\midrule

\texttt{THE HISTORY OF ANCIENT CIVILIZATIONS IS COMPLEX AND MULTIFACETED} &
\texttt{THE H1STORY 0F ANC1ENT CIVIL1ZATI0NS IS C0MPLEX AND MULT1FACETED} &
7 &
0.93 &
LD = 7 / 62 characters ($\approx 11\%$ error); minor substitutions, raw LD appears large; TRS penalizes incorrect content proportionally. \\

\addlinespace[3mm]

\texttt{THE HISTORY OF ANCIENT CIVILIZATIONS IS COMPLEX AND MULTIFACETED} &
\texttt{xQ9@!\#\%\textbackslash{}\^{}*()? \textgreater{} \textless{}\{\} [] \textbar{} \textasciitilde{} abcd1234567890} &
101 &
0.00 &
LD = 101 / 62 characters ($\approx 163\%$ error); complete corruption, raw LD grows unbounded; TRS heavily penalizes missing or incorrect content. \\

\addlinespace[3mm]

\texttt{HELLO WORLD} &
\texttt{HELLO} &
6 &
0.94 &
LD = 6 / 11 characters ($\approx 55\%$ error); recovered text is shorter; TRS penalizes missing content. \\

\addlinespace[3mm]

\texttt{HELLO} &
\texttt{HELLO ABCDEF} &
6 &
0.94 &
LD = 6 / 5 characters ($\approx 120\%$ error); recovered text is longer; TRS penalizes extra content appropriately. \\

\bottomrule
\end{tabular}
\end{table}

Each row in Table \ref{tab:trs_comparison} illustrates a different type of text error-minor substitutions, complete corruption, missing content or extra content and shows how raw Levenshtein Distance (LD) and the Text Recovery Score (TRS) evaluate these cases. TRS normalizes and clamps the edit distance producing a bounded score between $0$ and $1$ and proportionally penalizes missing or extra content providing a more interpretable measure than raw LD. This metric is consistent with previous work on normalized text similarity metrics which are bounded to $[0,1]$.

The second metric is a log-scaled Levenshtein-based similarity which applies a logarithmic transformation to the raw edit distance. This emphasizes small differences while compressing the penalty for large errors. The score lies strictly in $(0,1]$, with $1$ corresponding to a perfect match and values approaching $0$ for highly divergent texts:

\begin{equation}
    \text{LogLevSimilarity}(LD) = \frac{1}{1 + \log_{10}(1 + LD)}.
\end{equation}
\noindent
This score maps the raw edit distance to the interval $(0,1]$, with $1$ indicating a perfect match and values tending toward $0$ as the dissimilarity between the texts increases.

\begin{table}[H]
\centering
\setlength{\tabcolsep}{8pt}
\caption{Comparison of Raw Levenshtein Distance (LD) and Log-scaled Levenshtein Similarity (LLS).}
\label{tab:loglev_comparison}
\begin{tabular}{p{3.5cm} p{3.5cm} c c >{\raggedright\arraybackslash}p{4.6cm}}
\toprule
\textbf{Original Text} &
\textbf{Estimated Text} &
\textbf{LD} &
\textbf{LLS} &
\textbf{Error Analysis} \\
\midrule

\texttt{INTERNATIONALIZATION} &
\texttt{INTERNAT1ONALIZAT1ON} &
2 &
0.676 &
LD = 2 / 20 characters ($\approx 10\%$ error); minor substitutions penalized disproportionately in long text; LLD reflects relative error more appropriately. \\[2mm]

\addlinespace[3mm]

\texttt{CAT} &
\texttt{CUT} &
1 &
0.768 &
LD = 1 / 3 characters ($\approx 33\%$ error); short word appears ``better'' under raw LD; LLD accounts for proportional significance. \\[2mm]

\addlinespace[3mm]

\texttt{HELLO WORLD} &
\texttt{HELLO} &
6 &
0.765 &
LD = 6 / 11 characters ($\approx 55\%$ error); recovered text is shorter, raw LD underestimates proportional error; LLD heavily penalizes missing content. \\[2mm]

\addlinespace[3mm]

\texttt{HELLO} &
\texttt{HELLO ABCDEF} &
6 &
0.765 &
LD = 6 / 5 characters ($\approx 120\%$ error); recovered text is longer, raw LD underestimates proportional error; LLD penalizes extra content as an important factor. \\

\bottomrule
\end{tabular}
\end{table}

Table \ref{tab:loglev_comparison} compares Log-scaled Levenshtein Similarity (LLS) with raw Levenshtein Distance (LD) . LLS compresses small edit distances via a logarithmic scale transformation producing values bounded in $(0, 1]$. On the other side, LD simply counts minor substitutions from the original text without considering proportional significance or text length. Moreover, LLS provides a more interpretable measure of text recovery quality across texts of different lengths penalizing extra or missing content proportionally.

\subsection{Results \& Experiments}
Our method has been implemented in Python using TensorFlow \cite{tensorflow2015-whitepaper}, and all experiments were conducted on an NVIDIA GeForce RTX 2080 Ti GPU with 11,GB GDDR6 RAM and a 1350,MHz base clock. We use L-BFGS \cite{LBFGS} as the optimization algorithm and set the maximum number of epochs to $5000$ (as dictated by the behavior of the L-BFGS optimizer).

In this subsection, we present the results of our experiments on all three datasets (see Appendix\ref{sec:appendixA}, in particular Figures\ref{fig:DatasetAExemplars}, \ref{fig:DatasetBExemplars}, and \ref{fig:DatasetCExemplars}). The datasets are used as exemplars in MESA to guide the recovery of high-quality inscription images, and as ground-truth images for the training of DnCNN, ESRGAN, and DRUNET, from which synthetic ruined images (with scratches or noise) are generated. In all figures and tables below, the method annotations are as follows: \textit{TF} denotes a training-free method (no training required), \textit{TN} denotes a model trained on noisy textures, and \textit{TS} denotes a model trained on scratched textures.

\begin{figure}[H]
    \centering
    \includegraphics[width=0.9\textwidth,height=0.83\textheight,keepaspectratio]{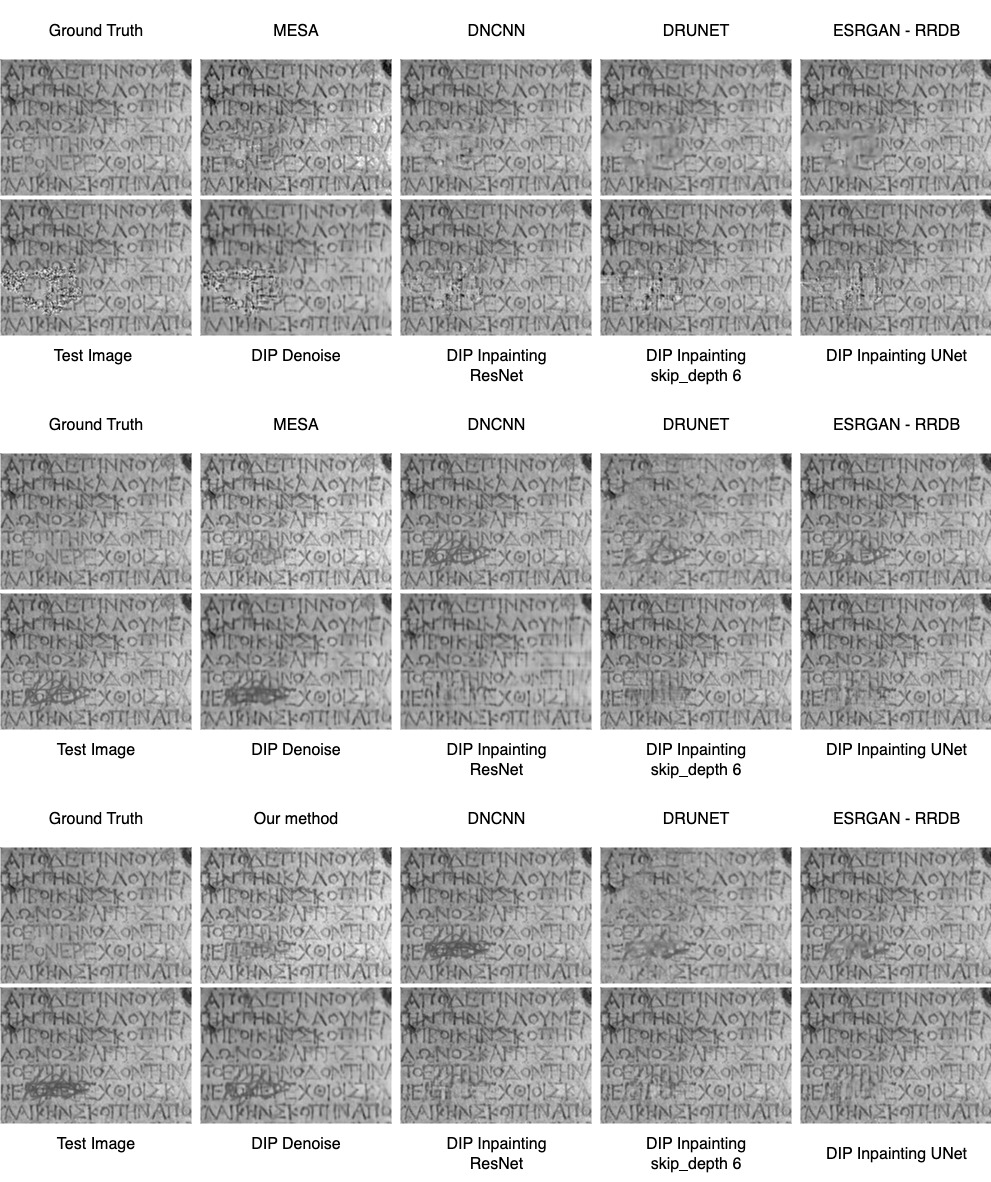}
    
    \caption{Comparison of our method's outputs per test image using Dataset A images as exemplars with other well-known methods.}
    
    \Description{}
    
    \label{fig:datasetAResultImages}
\end{figure}

\begin{figure}[H]
    \centering
    \includegraphics[width=0.9\textwidth,height=0.83\textheight,keepaspectratio]{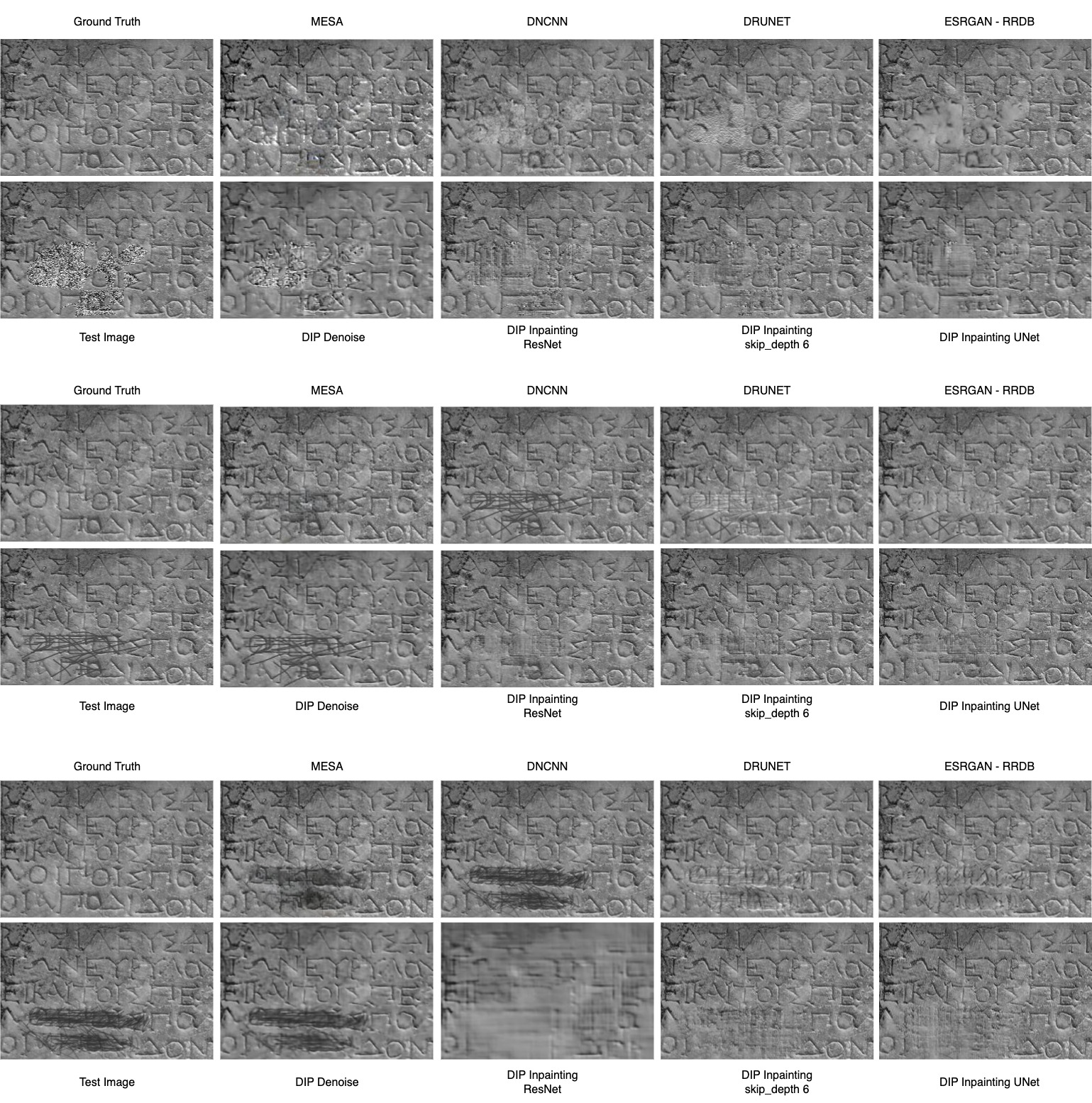}
    
    \caption{Comparison of our method's outputs per test image using Dataset B images as exemplars with other well-known methods.}
    
    \Description{}
    
    \label{fig:datasetBResultImages}
\end{figure}

\begin{figure}[H]
    \centering
    \includegraphics[width=0.9\textwidth,height=0.83\textheight,keepaspectratio]{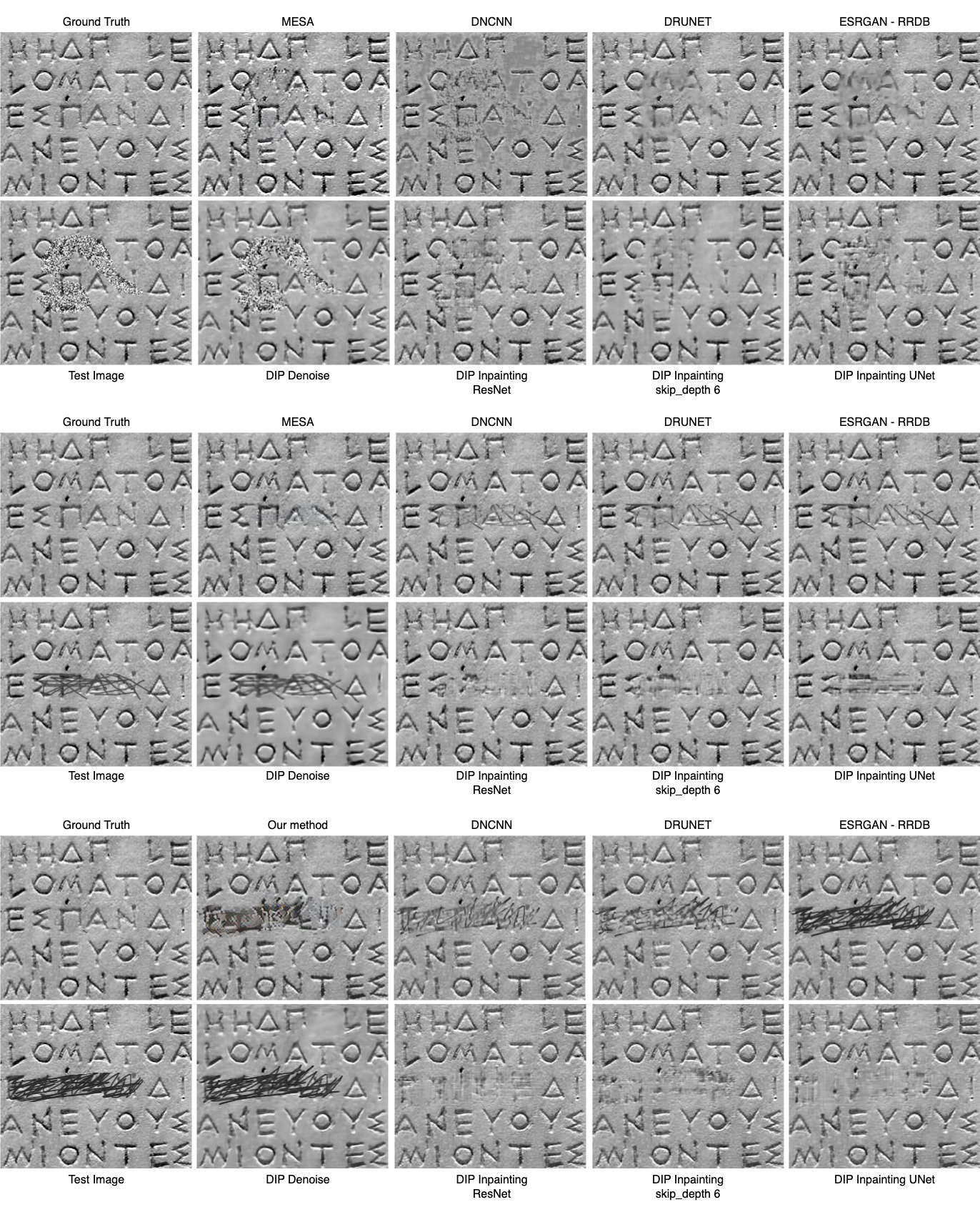}
    
    \caption{Comparison of our method's outputs per test image using Dataset C images as exemplars with other well-known methods.}
    
    \Description{}
    
    \label{fig:datasetCResultImages}
\end{figure}

Figures~\ref{fig:datasetAResultImages}, \ref{fig:datasetBResultImages}, and \ref{fig:datasetCResultImages} illustrate our experiments on recovering ruined inscription text from degradation caused either by noise (e.g., due to humidity or other environmental factors) or by scratches as an example of human-induced damage. In all three figures, it is evident that our method performs well in visually restoring the damaged text, whereas other methods either fail to produce visually acceptable results or, when they do, incur a high training cost due to the need for generating ruined textures. In these experiments, the training sets are constructed from images belonging to different datasets. This strategy improves generalization and reduces the cost of creating expensive training datasets by leveraging publicly available resources. 

However, in Appendix~\ref{sec:appendixA} we also report the results of an idealized training scenario, simulating the performance of methods that require training when ruined/ground-truth pairs are available from the same inscription or dataset. Among all baselines, Deep Image Prior (DIP) is the only method comparable to ours in terms of computational and data cost, since it does not require paired ruined and ground-truth images for training. For the remaining methods, we observe that text recovery is often less visually satisfactory than ours, or at best comparable in terms of legibility, while still requiring either synthetically generated ruined inscriptions —whose realism and coverage are difficult to guarantee— or large collections of real damaged samples, whose acquisition is typically impractical and costly.

In all Tables and Figures below it is important to note that we present the different metric evaluations for each separate Dataset as an average of input inscription images depicted above in all Figures \ref{fig:datasetAResultImages}, \ref{fig:datasetBResultImages} and \ref{fig:datasetCResultImages}.
 

\begin{table}[H]
\centering
\begin{tabular}{|l|c|c|c|c|}
\hline
\textbf{Method} & \textbf{Dataset A} & \textbf{Dataset B} & \textbf{Dataset C} & \textbf{Average} \\
\hline
DIP-DENOISE TF & 70.3333 & 37.0000 & 32.0000 & 46.4444 \\
DIP-INPAINTING TF & 70.2222 & 32.5556 & 32.6667 & 45.1481 \\
DNCNN TN & 63.3333 & 51.3333 & 30.3333 & 48.3333 \\
DNCNN TS & 68.6667 & 54.0000 & 34.0000 & 52.2222 \\
DRUNET TN & 67.6667 & 48.3333 & 37.3333 & 51.1111 \\
DRUNET TS & 86.3333 & 47.0000 & 31.3333 & 54.8889 \\
ESRGAN TN & 65.0000 & 55.6667 & 29.0000 & 49.8889 \\
ESRGAN TS & 65.0000 & 64.0000 & 34.0000 & 54.3333 \\
MESA TF & 68.3333 & 36.0000 & 26.6667 & 43.6667 \\
\hline
\end{tabular}
\caption{Levenshtein Distance per method across datasets and their averages.}
\label{tab:LevenshteinDistanceTable}
\end{table}

\begin{figure}[H]
    \centering
    
    \begin{subfigure}[b]{0.45\textwidth}
        \centering
        \includegraphics[width=\textwidth]{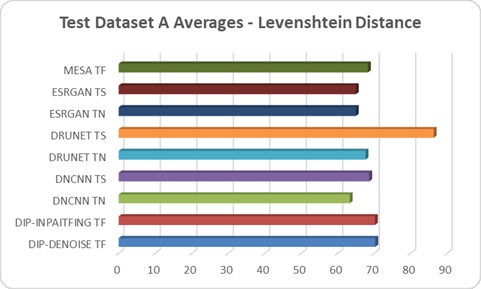}
        \caption{Dataset A}
    \end{subfigure}
    \hfill
    \begin{subfigure}[b]{0.45\textwidth}
        \centering
        \includegraphics[width=\textwidth]{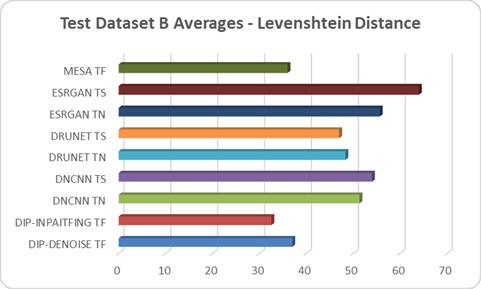}
        \caption{Dataset B}
    \end{subfigure}
    
    \vspace{1em} 

    \begin{subfigure}[b]{0.45\textwidth}
        \centering
        \includegraphics[width=\textwidth]{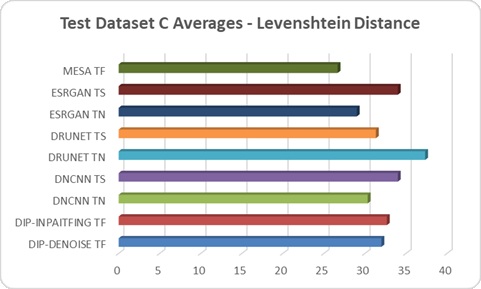}
        \caption{Dataset C}
    \end{subfigure}
    \hfill
    \begin{subfigure}[b]{0.45\textwidth}
        \centering
        \includegraphics[width=\textwidth]{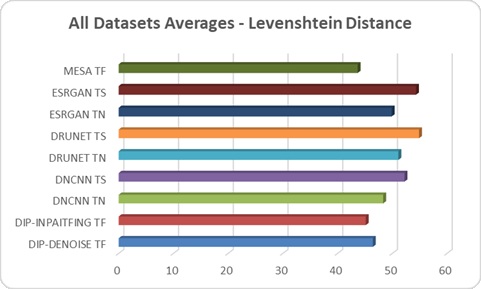}
        \caption{All Datasets' Averages}
    \end{subfigure}
    
    \caption{Levenshtein Distance evaluation}
    \Description{Levenshtein Distance evaluation}
    \label{fig:LevenshteinDistanceFigure}
\end{figure}

In Figure~\ref{fig:LevenshteinDistanceFigure} and Table~\ref{tab:LevenshteinDistanceTable}, we present the Levenshtein distance results, from which we observe that our method, on average, outperforms the baselines by requiring fewer operations to transform the recovered output into the ground-truth text. A lower distance indicates a better restoration quality. Both the ground-truth and recovered texts are obtained using Tesseract~\cite{tesseract}, and the same procedure is applied to all results presented in these figures.

\begin{table}[H]
\centering
\begin{tabular}{|l|c|c|c|c|}
\hline
\textbf{Method} & \textbf{Dataset A} & \textbf{Dataset B} & \textbf{Dataset C} & \textbf{Average} \\
\hline
DIP-DENOISE TF & 0.41 & 0.6967 & 0.7233 & 0.61 \\
DIP-INPAINTING TF & 0.4033 & 0.7378 & 0.7189 & 0.62 \\
DNCNN TN & 0.4467 & 0.58 & 0.7333 & 0.5867 \\
DNCNN TS & 0.4033 & 0.5767 & 0.7067 & 0.5622 \\
DRUNET TN & 0.4133 & 0.6033 & 0.7 & 0.5722 \\
DRUNET TS & 0.2767 & 0.6333 & 0.7333 & 0.5478 \\
ESRGAN TN & 0.44 & 0.58 & 0.7633 & 0.5944 \\
ESRGAN TS & 0.4567 & 0.4933 & 0.7167 & 0.5556 \\
MESA TF & 0.42 & 0.7033 & 0.77 & 0.6311 \\
\hline
\end{tabular}
\caption{Text Recovery Score per method across datasets and their averages.}
\label{tab:TRSResultsTable}
\end{table}

\begin{figure}[H]
    \centering
    
    \begin{subfigure}[b]{0.45\textwidth}
        \centering
        \includegraphics[width=\textwidth]{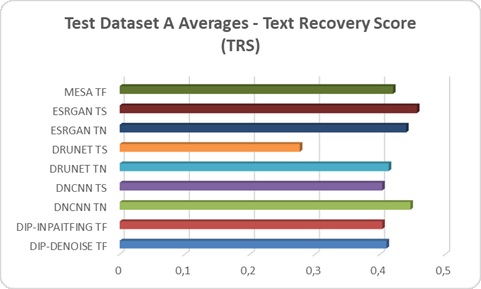}
        \caption{Dataset A}
    \end{subfigure}
    \hfill
    \begin{subfigure}[b]{0.45\textwidth}
        \centering
        \includegraphics[width=\textwidth]{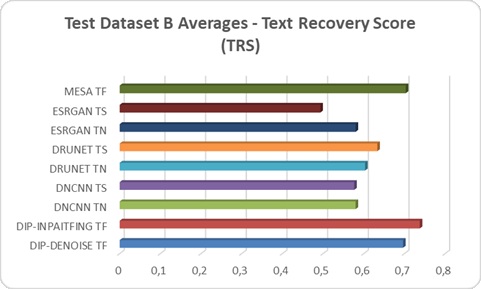}
        \caption{Dataset B}
    \end{subfigure}

    \vspace{1em} 

    \begin{subfigure}[b]{0.45\textwidth}
        \centering
        \includegraphics[width=\textwidth]{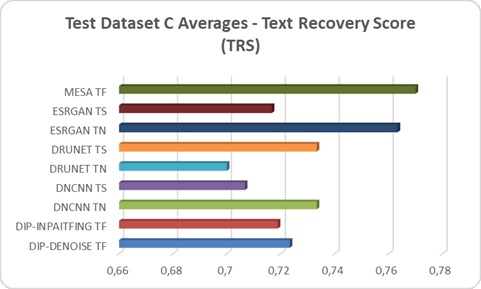}
        \caption{Dataset C}
    \end{subfigure}
    \hfill
    \begin{subfigure}[b]{0.45\textwidth}
        \centering
        \includegraphics[width=\textwidth]{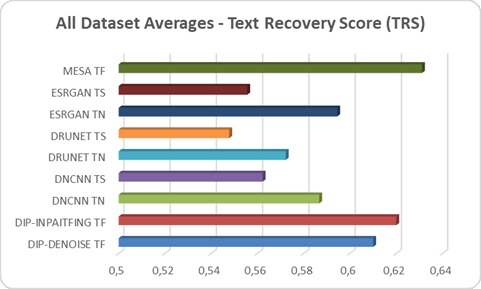}
        \caption{All Datasets' Averages}
    \end{subfigure}
    
    \caption{Text Recovery Score (TRS) evaluation}
    \Description{Text Recovery Score (TRS) evaluation}
    
    \label{fig:TRSFigure}
\end{figure}

Table \ref{tab:TRSResultsTable} and Figure \ref{fig:TRSFigure} demonstrate that our method  outperforms competing approaches in terms of text recovery, as quantified by the average Text Recovery Score introduced in Section~\ref{subsec:SubSectionMetrics}. 
Across all datasets, as well as in the aggregated dataset averages, the effectiveness of our method is consistently highlighted relative to the other methods.

\begin{table}[H]
\centering
\begin{tabular}{|l|c|c|c|c|}
\hline
\textbf{Method} & \textbf{Dataset A} & \textbf{Dataset B} & \textbf{Dataset C} & \textbf{Average} \\
\hline
DIP-DENOISE TF & 0.3507 & 0.3885 & 0.3971 & 0.3788 \\
DIP-INPAINTING TF & 0.3508 & 0.3973 & 0.3964 & 0.3815 \\
DNCNN TN & 0.3566 & 0.3679 & 0.4015 & 0.3753 \\
DNCNN TS & 0.3524 & 0.3688 & 0.3936 & 0.3716 \\
DRUNET TN & 0.3527 & 0.3768 & 0.3877 & 0.3724 \\
DRUNET TS & 0.3401 & 0.3768 & 0.3991 & 0.3720 \\
ESRGAN TN & 0.3550 & 0.3634 & 0.4043 & 0.3743 \\
ESRGAN TS & 0.3567 & 0.3725 & 0.3938 & 0.3743 \\
MESA TF & 0.3527 & 0.3897 & 0.4104 & 0.3843 \\
\hline
\end{tabular}
\caption{Log-scaled Levenshtein Similarity per method across datasets and their averages.}
\label{tab:logLevenshteinDistanceTable}
\end{table}

\begin{figure}[H]
    \centering
    
    \begin{subfigure}[b]{0.45\textwidth}
        \centering
        \includegraphics[width=\textwidth]{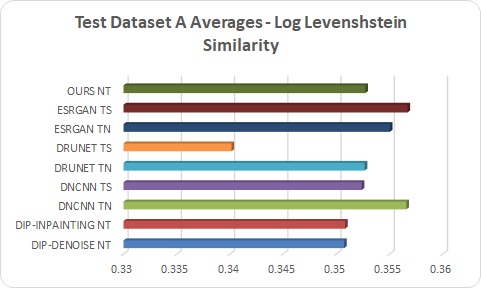}
        \caption{Dataset A}
    \end{subfigure}
    \hfill
    \begin{subfigure}[b]{0.45\textwidth}
        \centering
        \includegraphics[width=\textwidth]{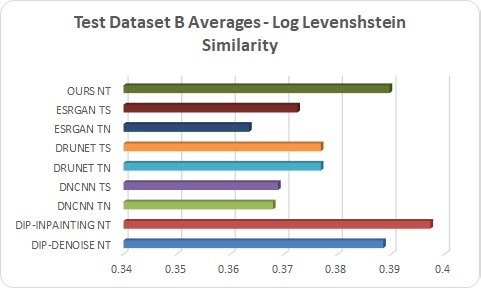}
        \caption{Dataset B}
    \end{subfigure}

    \vspace{1em} 

    \begin{subfigure}[b]{0.45\textwidth}
        \centering
        \includegraphics[width=\textwidth]{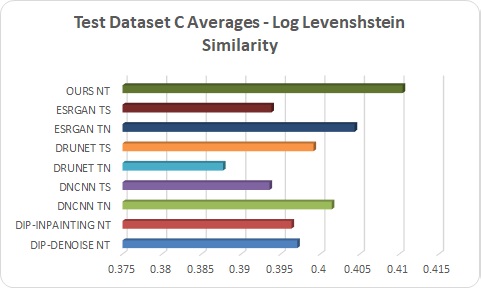}
        \caption{Dataset C}
    \end{subfigure}
    \hfill
    \begin{subfigure}[b]{0.45\textwidth}
        \centering
        \includegraphics[width=\textwidth]{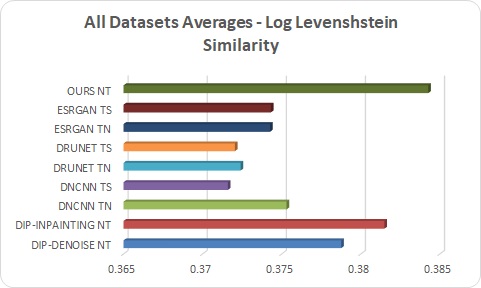}
        \caption{All Datasets' Averages}
    \end{subfigure}
    
    \caption{Log-scaled Levenshtein Similarity evaluation as described in \ref{subsec:SubSectionMetrics}}
    
    \Description{Log-scaled Levenshtein Similarity evaluation as described in \ref{subsec:SubSectionMetrics}}
    \label{fig:LogLevenshteinDistanceFigure}
\end{figure}

Figure \ref{fig:LogLevenshteinDistanceFigure} and Table \ref{tab:logLevenshteinDistanceTable} illustrate that our method surpasses the other methods in text recovery as measured by the Log Levenshtein Distance metric which is a newly proposed metric as presented in \ref{subsec:SubSectionMetrics}. This metric makes clear that our approach is better than no training needed counterparts (Deep Image Prior), but also from well-known networks utilized for restoring ruined inscription parts, which networks need training as is mentioned above. The higher the value the better the restoration quality.

\begin{table}[H]
\centering
\begin{tabular}{|l|c|c|c|c|}
\hline
\textbf{Method} & \textbf{Dataset A} & \textbf{Dataset B} & \textbf{Dataset C} & \textbf{Average} \\
\hline
DIP-DENOISE TF & 0.1652 & 0.2253 & 0.2426 & 0.2111 \\
DIP-INPAINTING TF & 0.0729 & 0.1751 & 0.1020 & 0.1166 \\
DNCNN TN & 0.0435 & 0.0927 & 0.1621 & 0.0994 \\
DNCNN TS & 0.0893 & 0.1606 & 0.1373 & 0.1291 \\
DRUNET TN & 0.1247 & 0.1196 & 0.0879 & 0.1108 \\
DRUNET TS & 0.2192 & 0.1334 & 0.1542 & 0.1690 \\
ESRGAN TN & 0.0495 & 0.0928 & 0.0871 & 0.0764 \\
ESRGAN TS & 0.0894 & 0.1187 & 0.1614 & 0.1231 \\
MESA TF & 0.0552 & 0.0993 & 0.0839 & 0.0795 \\
\hline
\end{tabular}
\caption{LPIPS Alex Network used per method across datasets and their averages.}
\label{tab:LPIPSAlexTable}
\end{table}

\begin{figure}[H]
    \centering
    
    \begin{subfigure}[b]{0.45\textwidth}
        \centering
        \includegraphics[width=\textwidth]{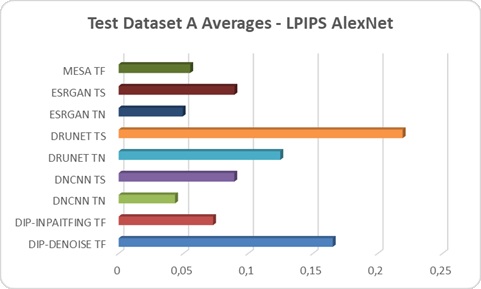}
        \caption{Dataset A}
    \end{subfigure}
    \hfill
    \begin{subfigure}[b]{0.45\textwidth}
        \centering
        \includegraphics[width=\textwidth]{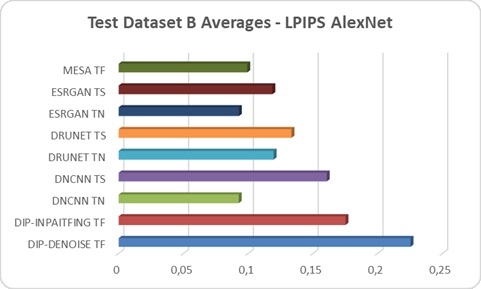}
        \caption{Dataset B}
    \end{subfigure}

    \vspace{1em} 

    \begin{subfigure}[b]{0.45\textwidth}
        \centering
        \includegraphics[width=\textwidth]{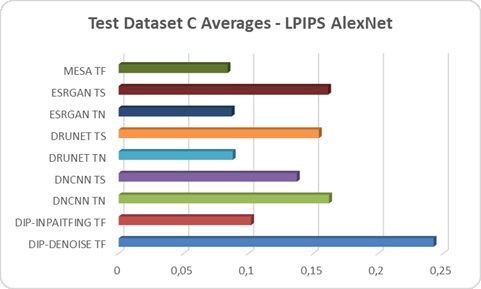}
        \caption{Dataset C}
    \end{subfigure}
    \hfill
    \begin{subfigure}[b]{0.45\textwidth}
        \centering
        \includegraphics[width=\textwidth]{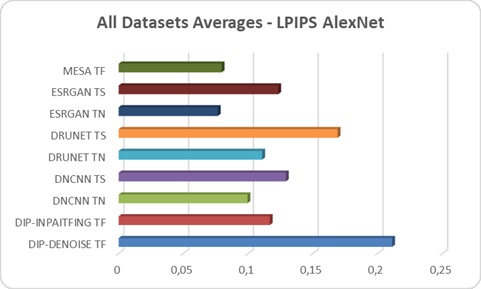}
        \caption{All Datasets' Averages}
    \end{subfigure}
    
    \caption{LPIPS evaluation with Alex network usage}
    \Description{LPIPS evaluation with Alex network usage}
    
    \label{fig:LPIPSAlexFigure}
\end{figure}


Figure \ref{fig:LPIPSAlexFigure} and Table \ref{tab:LPIPSAlexTable} highlight the effectiveness of our method with respect to a metric that more closely reflects human perception. LPIPS is a perceptual measure that leverages features from pre-trained CNNs to quantify the similarity between the feature representations of two images. Our method performs on par with ESRGAN on average (where we use only the generator component), while outperforming all other baselines. 

When instantiated with the AlexNet backbone, LPIPS captures how well the overall, large-scale structure of the recovered image matches the original, placing greater emphasis on shapes, strokes, and contours than on fine-grained texture differences. This makes it better suited for inscription restoration than the VGG-based LPIPS variant, whose deeper architecture is more sensitive to subtle texture and contrast patterns. Nevertheless, for completeness, next we also report results using the VGG-based LPIPS metric.



\begin{figure}[H]
    \centering
    
    \begin{subfigure}[b]{0.45\textwidth}
        \centering
        \includegraphics[width=\textwidth]{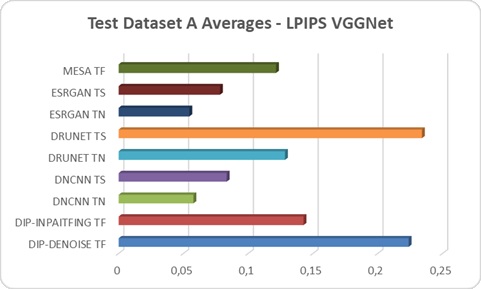}
        \caption{Dataset A}
    \end{subfigure}
    \hfill
    \begin{subfigure}[b]{0.45\textwidth}
        \centering
        \includegraphics[width=\textwidth]{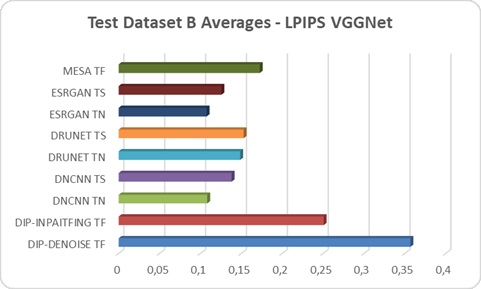}
        \caption{Dataset B}
    \end{subfigure}

    \vspace{1em} 

    \begin{subfigure}[b]{0.45\textwidth}
        \centering
        \includegraphics[width=\textwidth]{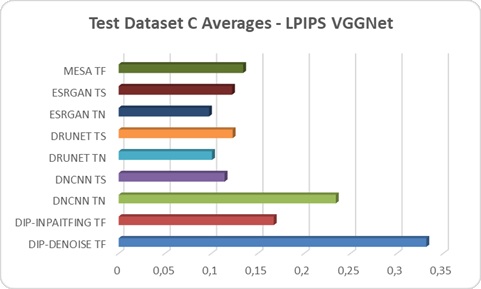}
        \caption{Dataset C}
    \end{subfigure}
    \hfill
    \begin{subfigure}[b]{0.45\textwidth}
        \centering
        \includegraphics[width=\textwidth]{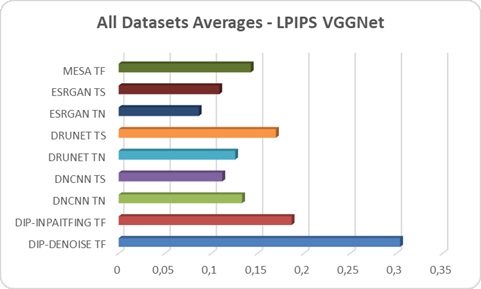}
        \caption{All Datasets' Averages}
    \end{subfigure}
    
    \caption{LPIPS evaluation with VGG network usage}
    \Description{LPIPS evaluation with VGG network usage}
    
    \label{fig:LPIPSVGGFigure}
\end{figure}

Table \ref{tab:LPIPSVGGTable} and Figure \ref{fig:LPIPSVGGFigure} report the performance of our method, which is substantially better than training-free alternatives such as Deep Image Prior variants. As noted above, LPIPS is a perceptual metric that leverages pre-trained CNN features to compare the feature spaces of two images; in this VGG-based variant, higher importance is assigned to fine-grained differences between the original and recovered images (with lower values indicating better similarity). This metric therefore penalizes minor artifacts that are not critical for letter and texture readability. 

Our primary objective is to recover the global structure and shapes of damaged letters and words, rather than to perfectly reproduce all low-level texture details, which lie somewhat outside the core scope of text restoration. Nonetheless, our method still achieves competitive VGG-based LPIPS scores, even for subtle details, and these could be further improved through multiple passes of our method, complementary techniques, or manual post-processing.

\begin{table}[H]
\centering
\begin{tabular}{|l|c|c|c|c|}
\hline
\textbf{Method} & \textbf{Dataset A} & \textbf{Dataset B} & \textbf{Dataset C} & \textbf{Average} \\
\hline
DIP-DENOISE TF & 0.2238 & 0.3568 & 0.3318 & 0.3041 \\
DIP-INPAINTING TF & 0.1427 & 0.2507 & 0.1672 & 0.1869 \\
DNCNN TN & 0.0577 & 0.1083 & 0.2342 & 0.1334 \\
DNCNN TS & 0.0834 & 0.1382 & 0.1142 & 0.1119 \\
DRUNET TN & 0.1284 & 0.1486 & 0.1007 & 0.1259 \\
DRUNET TS & 0.2341 & 0.1528 & 0.1229 & 0.1699 \\
ESRGAN TN & 0.0545 & 0.1076 & 0.0974 & 0.0865 \\
ESRGAN TS & 0.0782 & 0.1257 & 0.1222 & 0.1087 \\
MESA TF & 0.1214 & 0.1725 & 0.1343 & 0.1428 \\
\hline
\end{tabular}
\caption{LPIPS VGG Network used per method across datasets and their averages.}
\label{tab:LPIPSVGGTable}
\end{table}

\begin{table}[H]
\centering
\begin{tabular}{|l|c|c|c|c|}
\hline
\textbf{Method} & \textbf{Dataset A} & \textbf{Dataset B} & \textbf{Dataset C} & \textbf{Average} \\
\hline
DIP-DENOISE TF & 0.8269 & 0.7059 & 0.6922 & 0.7416 \\
DIP-INPAINTING TF & 0.8834 & 0.7548 & 0.8212 & 0.8198 \\
DNCNN TN & 0.9590 & 0.9066 & 0.6169 & 0.8275 \\
DNCNN TS & 0.9458 & 0.8855 & 0.9094 & 0.9136 \\
DRUNET TN & 0.8393 & 0.8579 & 0.9182 & 0.8718 \\
DRUNET TS & 0.8174 & 0.8703 & 0.9040 & 0.8639 \\
ESRGAN TN & 0.9607 & 0.9133 & 0.9203 & 0.9314 \\
ESRGAN TS & 0.9482 & 0.8938 & 0.9071 & 0.9164 \\
MESA TF & 0.9046 & 0.8393 & 0.8571 & 0.8670 \\
\hline
\end{tabular}
\caption{SSIM per method across datasets and their averages.}
\label{tab:SSIMResultsTable}
\end{table}

\begin{figure}[H]
    \centering
    
    \begin{subfigure}[b]{0.45\textwidth}
        \centering
        \includegraphics[width=\textwidth]{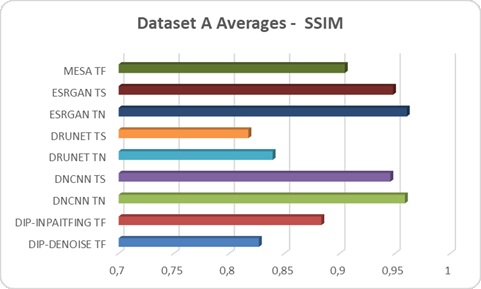}
        \caption{Dataset A}
    \end{subfigure}
    \hfill
    \begin{subfigure}[b]{0.45\textwidth}
        \centering
        \includegraphics[width=\textwidth]{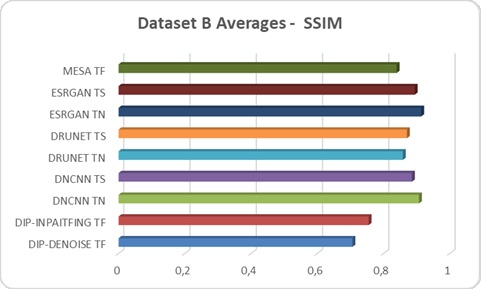}
        \caption{Dataset B}
    \end{subfigure}

    \vspace{1em} 

    \begin{subfigure}[b]{0.45\textwidth}
        \centering
        \includegraphics[width=\textwidth]{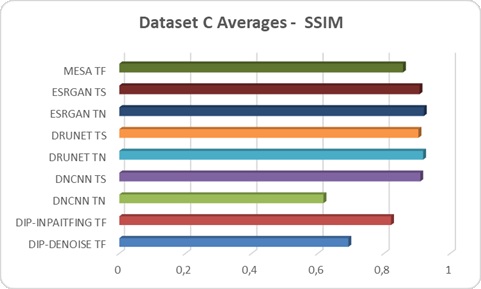}
        \caption{Dataset C}
    \end{subfigure}
    \hfill
    \begin{subfigure}[b]{0.45\textwidth}
        \centering
        \includegraphics[width=\textwidth]{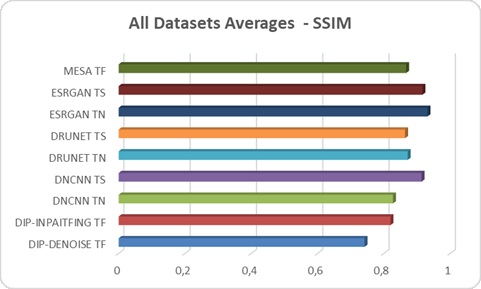}
        \caption{All Datasets' Averages}
    \end{subfigure}
    
    \caption{SSIM evaluation}
    \Description{SSIM evaluation}

    \label{fig:SSIMFigure}
\end{figure}

Table\ref{tab:SSIMResultsTable} and Figure\ref{fig:SSIMFigure} report the Structural Similarity Index (SSIM) scores. Although SSIM emphasizes local structural fidelity and is sensitive to pixel-level luminance and contrast, our method is not explicitly designed to optimize these low-level characteristics. This explains why, on average, it exhibits slightly lower SSIM performance compared to some training-based methods (ESRGAN, DnCNN, DRUNET), while still outperforming low-cost approaches such as Deep Image Prior (DIP), which require neither training nor dedicated datasets, as shown in both the figure and the table.

\begin{table}[H]
\centering
\begin{tabular}{|l|c|c|c|c|}
\hline
\textbf{Method} & \textbf{Dataset A} & \textbf{Dataset B} & \textbf{Dataset C} & \textbf{Averages} \\
\hline
DIP-DENOISE TF & 29.6638 & 29.9005 & 29.7345 & 29.7663 \\
DIP-INPAINTING TF & 31.2086 & 31.1740 & 31.3163 & 31.2330 \\
DNCNN TN & 38.4847 & 36.8845 & 31.3225 & 35.5639 \\
DNCNN TS & 38.3920 & 36.8139 & 36.0718 & 37.0926 \\
DRUNET TN & 30.8336 & 32.4036 & 36.2790 & 33.1720 \\
DRUNET TS & 31.0895 & 33.4126 & 35.7140 & 33.4054 \\
ESRGAN TN & 39.3524 & 37.5612 & 36.0042 & 37.6393 \\
ESRGAN TS & 39.1888 & 36.8152 & 36.6121 & 37.5387 \\
MESA TF & 28.9296 & 30.9083 & 31.2342 & 30.3574 \\
\hline
\end{tabular}
\caption{PSNR per method across datasets and their averages.}
\label{tab:PSNRResultsTable}
\end{table}

\begin{figure}[H]
    \centering
    
    \begin{subfigure}[b]{0.45\textwidth}
        \centering
        \includegraphics[width=\textwidth]{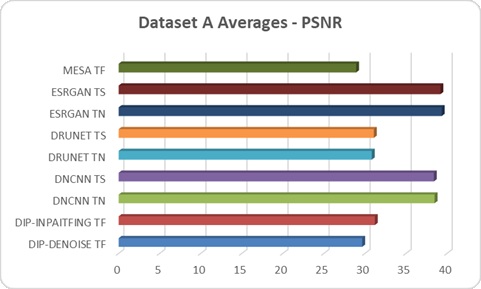}
        \caption{Dataset A}
    \end{subfigure}
    \hfill
    \begin{subfigure}[b]{0.45\textwidth}
        \centering
        \includegraphics[width=\textwidth]{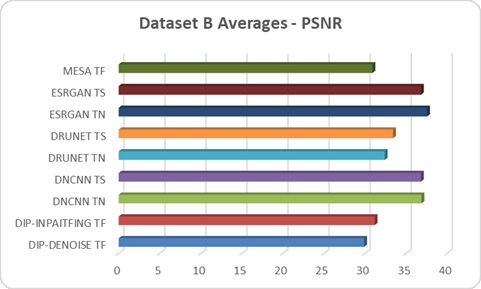}
        \caption{Dataset B}
    \end{subfigure}

    \vspace{1em} 

    \begin{subfigure}[b]{0.45\textwidth}
        \centering
        \includegraphics[width=\textwidth]{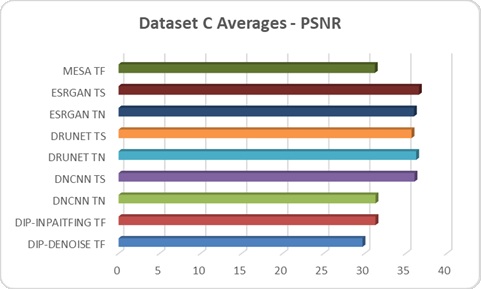}
        \caption{Dataset C}
    \end{subfigure}
    \hfill
    \begin{subfigure}[b]{0.45\textwidth}
        \centering
        \includegraphics[width=\textwidth]{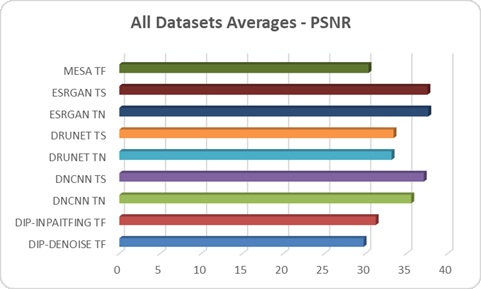}
        \caption{All Datasets' Averages}
    \end{subfigure}
    
    \caption{PSNR evaluation}
    \Description{PSNR evaluation}

    \label{fig:PSNRFigure}
\end{figure}

Table \ref{tab:PSNRResultsTable} and Figure \ref{fig:PSNRFigure} show the Peak Signal-to-Noise Ratio (PSNR) values achieved by the different methods. PSNR measures pixel-wise similarity between a reconstructed image and a reference (ground-truth) image, which is not the primary objective of our approach, as we focus on inscription text recovery rather than exact pixel reconstruction. Nonetheless, we report PSNR for completeness, since it is one of the most widely used metrics in image processing and computer vision.

Regenerate
Copy
Good response
Bad response

%% file: Sections/Conclusion.tex
\section{Discussion, Conclusions \& Future Work}

In this work, we introduced \emph{MESA}, a training-free, multi-exemplar, style-aware method for the restoration of degraded inscriptions. Unlike conventional supervised approaches such as DRUNET, DnCNN, and ESRGAN, our method does not require paired ruined/clean datasets or extensive pretraining. Instead, it leverages clean exemplars from the same or similar inscription styles to guide the reconstruction process, making it particularly suitable for cultural heritage scenarios where annotated data are scarce or prohibitively expensive to obtain. It also offers a competitive alternative to training-free baselines such as Deep Image Prior (DIP), while providing superior performance both visually and according to multiple quantitative metrics.

A key characteristic and potential limitation of MESA, despite its material independence, lies in its reliance on the availability of well-preserved and stylistically relevant exemplar inscriptions. The core mechanism of our approach, which leverages Gram matrices and minimizes Mean-Squared Displacement (MSD) for feature matching, is designed to capture and transfer texture, style, and stroke structure from these exemplars to the damaged input. This exemplar-driven approach is a strength in specific cultural heritage contexts, especially when dealing with multiple fragments from the same monument or a collection of inscriptions with a consistent aesthetic. In such scenarios, MESA excels because it has rich, relevant information to guide the reconstruction process, ensuring visual and stylistic coherence.

We evaluated MESA on three inscription datasets containing two major degradation types: human-induced scratches and noise due to environmental factors. To assess both visual quality and textual fidelity, we combined standard image-based metrics (PSNR, SSIM, LPIPS) with two newly proposed text-oriented measures: the Text Recovery Score (TRS) and a log-scaled Levenshtein-based similarity. These metrics jointly capture the trade-off between perceptual image quality and accurate letter and word reconstruction. Across all datasets and degradation types, our method consistently outperformed training-free baselines and matched or surpassed strong supervised competitors, while avoiding the need for costly paired training data or large synthetic datasets of damaged inscriptions.

Beyond empirical performance, our evaluation framework contributes two complementary, task-aware metrics that are better aligned with the goals of inscription restoration than generic image similarity measures alone. These metrics can be reused to assess future methods where text legibility and correctness are primary objectives.

Regarding generalizability across diverse script types, the reliance of MESA on VGG19 convolutional features for general visual attributes provides some inherent flexibility. However, the proposed character-scale weighting scheme, which utilizes OCR (Tesseract) for letter width analysis, introduces a script-dependent component. While Tesseract is highly effective for Latin and Greek alphabets (which are common in ancient inscriptions), its performance and the meaningfulness of its 'letter width' output could be severely limited for radically different writing systems, such as hieroglyphs, cuneiform, or ideographic scripts. These scripts have distinct structural characteristics that might not be accurately captured by an OCR system primarily trained on alphabetic languages, thus potentially impacting the layer weighting effectiveness.

Therefore, while the material independence of MESA is a significant advantage, its practical application is most suitable for contexts where a pool of stylistically consistent and adequately preserved exemplars can be accessed. Future work could explore more advanced, script-agnostic feature learning, or adaptive mechanisms that can learn relevant style cues from fewer, or even single, exemplars, potentially incorporating more sophisticated representations of graphical elements beyond simple character widths to enhance generalizability to highly diverse and unique inscription types.

Future work includes extending MESA to handle a wider variety of scripts and writing media, integrating explicit language models or epigraphic priors to further improve text plausibility, and exploring semi-automatic or interactive workflows in which experts can guide or refine the restoration process. We also plan to investigate more efficient optimization schemes and multi-scale formulations to reduce computation time while preserving the quality of the recovered inscriptions. Moreover, we could further investigate whether alternative conventional network architectures (beyond VGG19) which can provide improved quality or even design a new network specifically tailored to inscription restoration.